\documentclass[sn-mathphys]{sn-jnl}%

\normalbaroutside
\usepackage{todonotes}
\usepackage{listings}
\usepackage{xcolor}
\usepackage{comment}

\definecolor{codegreen}{rgb}{0,0.6,0}
\definecolor{codegray}{rgb}{0.5,0.5,0.5}
\definecolor{codepurple}{rgb}{0.58,0,0.82}
\definecolor{backcolour}{rgb}{0.95,0.95,0.92}

\lstdefinestyle{mystyle}{
    backgroundcolor=\color{backcolour},   
    commentstyle=\color{codegreen},
    keywordstyle=\color{magenta},
    numberstyle=\tiny\color{codegray},
    stringstyle=\color{codepurple},
    basicstyle=\ttfamily\footnotesize,
    breakatwhitespace=false,         
    breaklines=true,                 
    captionpos=b,                    
    keepspaces=true,                 
    numbers=left,                    
    numbersep=5pt,                  
    showspaces=false,                
    showstringspaces=false,
    showtabs=false,                  
    tabsize=2
}

\usepackage{upgreek}
\usepackage{caption}
\usepackage{subcaption}
\usepackage{tabularx}
\usepackage{bbding}
\usepackage{color,soul}
\usepackage{siunitx}
\usepackage{cleveref}

\jyear{2022}%
\theoremstyle{thmstyleone}%

\theoremstyle{thmstyletwo}%

\theoremstyle{thmstylethree}%

\DeclareMathOperator*{\argmin}{arg\,min}

\newcommand{\total}{\mathrm d}

\newcommand{\tx}[1]{\text{#1}}

\begin{document}

\title[Dis-AE Domain Generalisation]{Dis-AE: Multi-domain \& Multi-task Generalisation on Real-World Clinical Data}

\author[1]{\fnm{Daniel} \sur{Kreuter}}%
\equalcont{These authors contributed equally to this work.}

\author[1]{\fnm{Samuel} \sur{Tull}}%
\equalcont{These authors contributed equally to this work.}

\author[1]{\fnm{Julian} \sur{Gilbey}}

\author[2]{\fnm{Jacobus} \sur{Preller}}

\author[3]{\fnm{BloodCounts!} \sur{Consortium}}%

\author[4]{\fnm{John A.D.} \sur{Aston}}

\author[5]{\fnm{James H.F.} \sur{Rudd}}

\author[6]{\fnm{Suthesh} \sur{Sivapalaratnam}}

\author[1]{\fnm{Carola-Bibiane} \sur{Sch\"onlieb}}

\author[7,8]{\fnm{Nicholas} \sur{Gleadall}}

\author*[1]{\fnm{Michael} \sur{Roberts}}\email{mr808@cam.ac.uk}

\affil[1]{\orgdiv{Department of Applied Mathematics and Theoretical Physics}, \orgname{University of Cambridge}, \orgaddress{\street{Wilberforce Road}, \city{Cambridge}, \postcode{CB3 0WA}, \country{United Kingdom}}}

\affil[2]{\orgdiv{Addenbrooke's Hospital}, \orgaddress{\street{Cambridge University Hospitals}, \city{Cambridge}, \postcode{CB2 0QQ}, \country{United Kingdom}}}

\affil[3]{\orgdiv{A list of authors and their affiliations appears at the end of the paper}}

\affil[4]{\orgdiv{Department of Pure Mathematics and Mathematical Statistics}, \orgname{University of Cambridge}, \orgaddress{\street{Wilberforce Road}, \city{Cambridge}, \postcode{CB3 0WA}, \country{United Kingdom}}}

\affil[5]{\orgdiv{Department of Medicine}, \orgname{University of Cambridge, \city{Cambridge}, \country{UK}}}

\affil[6]{\orgname{Barts Health NHS Trust, \city{London}, \country{UK}}}

\affil[7]{\orgdiv{Department of Haematology}, \orgname{University of Cambridge}, \orgaddress{\street{Puddicombe Way, Cambridge Biomedical Campus}, \city{Cambridge}, \postcode{CB2 0AW}, \country{United Kingdom}}}

\affil[8]{\orgdiv{NHSBT}, \orgaddress{\street{Long Road}, \city{Cambridge}, \postcode{CB2 0PT}, \country{United Kingdom}}}

\abstract{
Clinical data is often affected by clinically irrelevant factors such as discrepancies between measurement devices or differing processing methods between sites. In the field of machine learning (ML), these factors are known as \textit{domains} and the distribution differences they cause in the data are known as \textit{domain shifts}. ML models trained using data from one domain often perform poorly when applied to data from another domain, potentially leading to wrong predictions. As such, developing machine learning models that can generalise well across multiple domains is a challenging yet essential task in the successful application of ML in clinical practice. In this paper, we propose a novel disentangled autoencoder (Dis-AE) neural network architecture that can learn domain-invariant data representations for multi-label classification of medical measurements even when the data is influenced by multiple interacting domain shifts at once.
The model utilises adversarial training to produce data representations from which the domain can no longer be determined.
We evaluate the model's domain generalisation capabilities on synthetic datasets and full blood count (FBC) data from blood donors as well as primary and secondary care patients, showing that Dis-AE improves model generalisation on multiple domains simultaneously while preserving clinically relevant information.
}


\maketitle
\section{Introduction}\label{sec:intro}

Machine learning has promised to revolutionise healthcare for several years \cite{sadegh-zadehDubioProAegro1990, handelmanEDoctorMachineLearning2018}. Moreover, while there is an extensive literature describing high-performing machine learning models trained on immaculate benchmark datasets \cite{milletariVNetFullyConvolutional2016, hanMADGANUnsupervisedMedical2021, caoSwinUnetUnetlikePure2021}, 
such promising approaches rarely make it into clinical practice \cite{teodoridisWhyAIAdoption2022}. Often, this is because of an unexpected drop in performance when deploying the model on unseen test data due to \textit{domain shift} \cite{rechtImageNetClassifiersGeneralize2019, moreno-torresUnifyingViewDataset2012}, i.e. there is a change in the data distribution between the dataset a model is trained on (\textit{source} data) and that which it is deployed against (\textit{target} data). Most common machine learning algorithms rely on an assumption that the source and target data are independent and identically distributed (i.i.d.) \cite{bishopPatternRecognitionMachine2006}. However, with domain shift, this assumption no longer holds, and model performance can be significantly affected. 
For medical datasets, domain shift is widespread, resulting from differences in equipment and clinical practice between sites \cite{liuSemisupervisedMetalearningDisentanglement2021, taoDeepLearningBased2019, thiagarajanUnderstandingBehaviorClinical2019, rolandDomainShiftsMachine2022}, and models are vulnerable to associating clinically irrelevant features specific to the domain with their predictions, known as \textit{shortcut learning}~\cite{geirhosShortcutLearningDeep2020}, which may lead to poor performance on target data.

For most medical applications, target data is rarely available prior to real-time deployment; thus, a \textit{domain adaptation} approach, where pre-trained models are fine-tuned on data from the target distribution is not feasible. \textit{Domain generalisation} techniques that focus on mitigating domain shift and improving model performance on unseen target data \cite{zhouDomainGeneralizationSurvey2022} are more practical approaches. Various techniques have been developed in recent years: \textit{data augmentation}~\cite{otaloraStainingInvariantFeatures2019, zhouLearningGenerateNovel2020}, where pre-processing is applied to source data in hopes of increasing data diversity; \textit{domain alignment}~\cite{liDomainGeneralizationMedical2020, huDomainGeneralizationMultidomain2020}, where domain shift is minimised by appropriate feature transformations which re-align the data before model training, and \textit{ensemble learning}~\cite{dingDeepDomainGeneralization2018, xuExploitingLowRankStructure2014, wangDoFEDomainOrientedFeature2020, nozzaDeepLearningEnsemble2016}, where an average of multiple models trained on different source distributions is taken. In medical datasets, particularly for tabular data, the extent and effect of unseen domains is often unknown, making alignment and augmentation impractical. For on-site deployment on non-specialised hardware, ensemble learning can often prove too expensive in terms of memory or computational time \cite{bifetNewEnsembleMethods2009}, particularly for deep neural network models.

\begin{figure}[h]
\centering
\includegraphics[width=\linewidth]{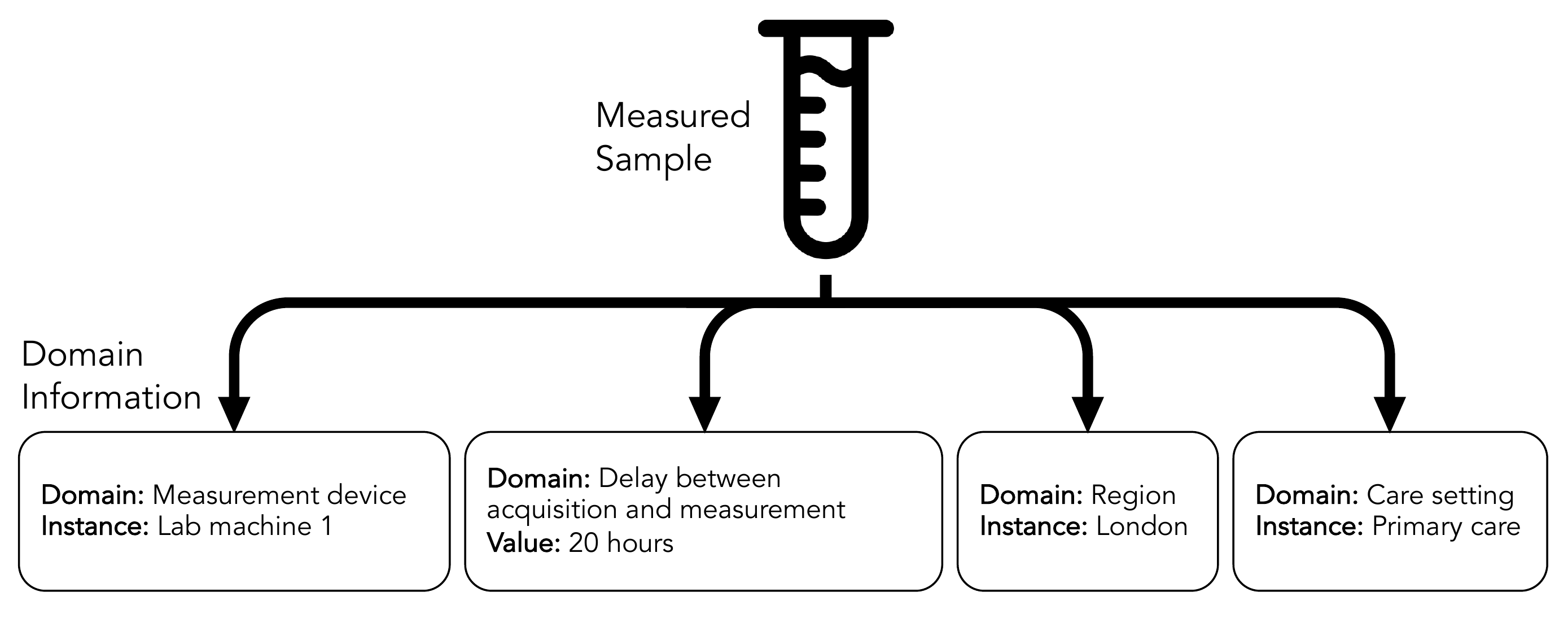}
\caption{\label{fig:domain_subdomain}
Schematic displaying the proposed domain-instance grouping. As an example, labels for medical data are shown. Notably, domains can either be characterised by continuous values or be made up of categorical instances.}
\end{figure}

Furthermore, in the sizeable domain generalisation literature, there is very little research into the problem of multiple interacting domains within any given dataset. In Figure~\ref{fig:domain_subdomain}, we indicate such a separation for some example domain shifts we observe in real-world datasets. 
Authors in the literature often refer to "domains" based solely on sample origin, rather than considering actual distribution-influencing effects causing domain shifts.

In this paper, we propose a deep learning model architecture based on feature disentanglement \cite{pmlr-v37-ganin15} to perform scalable domain generalisation for multiple interacting domains on tabular medical data. Such methods use multi-task learning to learn features useful for the desired task but penalise for learning features that can identify the sample's domain label. Importantly, no prior knowledge of the expression of the domain shift is required. Our novel approach allows for multiple domains and tasks, both continuous and categorical, creating a truly disentangled embedding that can be used for multiple classification tasks. We test the model on both synthetic and real-life clinical datasets, with the goal of retaining high performance on data not used in training.
The proposed architecture can be easily applied to other tabular data or images by adapting the corresponding model components.\\
Our novel contributions presented in this paper are:

\begin{itemize}
    \item We introduce a domain-instance grouping that goes beyond the current domain generalisation literature by identifying and separating effects that cause domain shift in the data (e.g. Figure~\ref{fig:domain_subdomain}).
    \item We propose a novel disentangled autoencoder (Dis-AE) neural network model architecture to find a domain-agnostic lower-dimensional representation of the data. Dis-AE is easily scalable to multiple domains and classification tasks.
    \item We conduct experiments of Dis-AE, comparing it to a regular autoencoder on synthetic and real-life clinical data. We show that Dis-AE achieves high domain generalisation performance when using the same width and depth as the regular autoencoder.
\end{itemize}

\section{Materials \& Methods}

\subsection*{Datasets}\label{sec:datasets}

We evaluate our model's domain generalisation performance on four synthetic datasets, data from two clinical studies, and hospital patient data.\\

\begin{figure}[h]
\centering
\includegraphics[width=\linewidth]{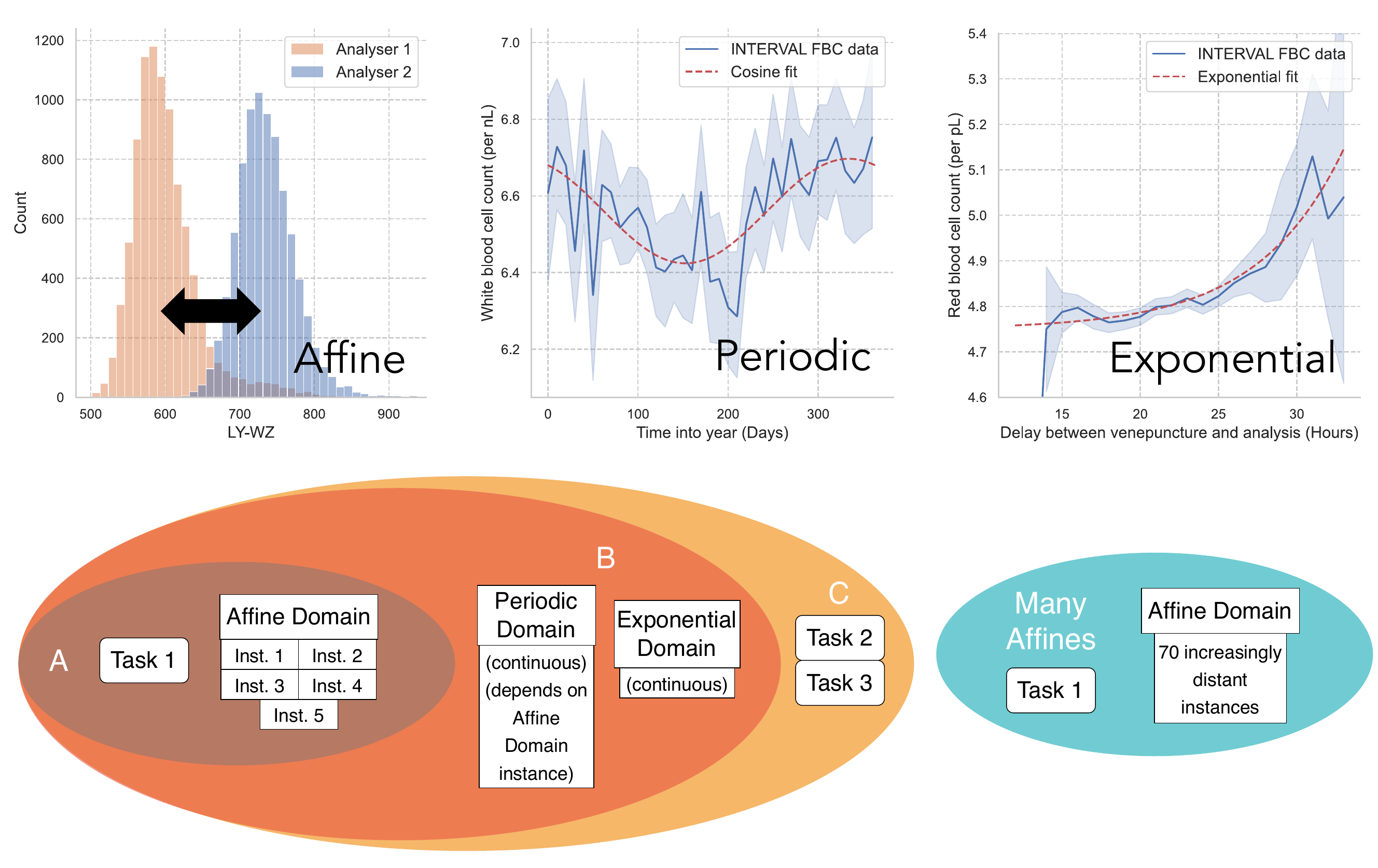}
\caption{\label{fig:datasets_schematic}
Top: Examples of domain shifts found in real-world full blood count data from the INTERVAL trial. These shifts inspired the three transformations used in the synthetic datasets. (LY-WZ: Forward-scattered light distribution width of the LYMPH area on the white cell differential fluorescence scattergram. In the blue line plots, the line and shaded area show the mean and \SI{95}{\percent} confidence interval of the data.)\\
Bottom: Schematic describing the synthetic datasets and their task and domain properties.
}
\end{figure}

\bmhead{Synthetic datasets}
The synthetic datasets \textbf{A}, \textbf{B}, and \textbf{C} represent a classification problem with increasing difficulty. Dataset A includes one task and one domain; dataset B adds multiple dependent continuous domains; dataset C adds two additional tasks (see Figure~\ref{fig:datasets_schematic}). The data was first generated using \textit{scikit-learn}'s~\cite{scikit-learn} \texttt{make\_classification} (A, B) and {\texttt{make\_multilabel\_classification}} (C) functions. Afterwards, we added the artificial domains using simple transformations on some features of the data.
Dataset A and Dataset B consist of \num{13000} samples each, focusing on a single task with two balanced classes. Dataset C, on the other hand, comprises \num{26000} samples involving three distinct tasks, each with two classes. While Task 2 maintains class balance, Task 1 and Task 3 exhibit imbalanced classes with a ratio of $5:2$.
To simulate domain shifts, we employed five affine instances in our datasets. These instances were created with a ratio of $5:5:1:1:1$, ensuring that the majority of samples belong to the two instances used as source data. Additionally, other domain shifts in Dataset B and Dataset C were uniformly distributed across all samples.\\
To explore the generalisation limits of our model, we used the same procedure to make the \textbf{Many Affines} dataset, which includes a large number of instances on one domain and a single task. The dataset contains a large-scale collection of \num{500000} samples, focusing on a single task with two balanced classes. Notably, these samples are evenly distributed among 70 distinct affine instances, offering a diverse range of transformation patterns for robust analysis.\\
The synthetic datasets and the applied domain transformations are summarised in Figure~\ref{fig:datasets_schematic} (bottom) and Appendix~\ref{sec:A_DatasetsGeneration}.

\bmhead{Clinical datasets}
We conduct experiments on multiple clinical datasets of haematological measurements.
Specifically, we conduct experiments on features corresponding to standard high-level full blood count (HL-FBC) measurements, as well as further research and clustering parameters recorded by the haematology analyser machine during a full blood count. We refer to this data with additional features as \textit{rich} full blood count (R-FBC) data. The parameters are described in Appendix~\ref{sec:A_FBC}.\\
The datasets from \textbf{INTERVAL}~\cite{angelantonioEfficiencySafetyVarying2017} and \textbf{COMPARE}~\cite{bellComparisonFourMethods2020}
contain R-FBC data from their respective clinical trials. These real-world datasets show natural domain shifts that occur when collecting R-FBC data in clinical practice (some shown in Figure~\ref{fig:datasets_schematic}). The studies collected samples from \num{48460} and \num{29029} donors, respectively. After pre-filtering (see Appendix~\ref{sec:A_FBC_preprocess}), we are left with \num{15438} and \num{12064} R-FBC samples, respectively.
Sysmex XN-2000 haematology analysers were used for both blood donor studies, with two separate machines used for INTERVAL’s analysis. The shift between analysers within the studies is strongly pronounced, as seen in Figure~\ref{fig:datasets_schematic}.
Each measurement includes the times of both the donation appointment and the subsequent analysis in the corresponding haematology analyser. Additional information, such as the delay between venepuncture and measurement as well as the date of donation, can be determined from these times. Therefore, we consider the analyser, the delay between venepuncture and measurement, and time into the year (seasonal shifts) as our three domains of interest which result in a shift in the data but should not have an influence on our classification tasks. The analyser domain contains three instances since two Sysmex XN-2000 analysers were used in INTERVAL and one in COMPARE, while continuous time values represent the other two domains. As classification tasks, we selected subject sex, BMI bracket, and age bracket. 
While arguably lacking practical clinical value, we mainly chose these classification objectives to illustrate our proposed method's performance and because they are clearly definable. The sample distributions for domains and tasks are displayed in Appendix~\ref{sec:A_FBC_preprocess}.\\
The dataset from \textbf{Cambridge University Hospitals (CUH)}~\cite{EpiCovVersion1a} contains HL-FBC data from real-life clinical practice, including primary and secondary care data from \num{438483} patients between 2016 and 2022. After pre-filtering (see Appendix~\ref{sec:A_EpiCov_FBC_preprocess}), we are left with \num{1036709} HL-FBC samples from \num{376056} patients. The data at CUH was recorded on 5 ADVIA 2120 analysers which were gradually replaced with ADVIA 2120i machines throughout 2021 and 2022. Since one analyser at CUH is used exclusively for emergency cases, we chose to exclude all samples from the emergency 2120 and 2120i machines, leaving 8 analyser instances. Here too, the distribution difference between analysers is significant (Kolmogorov-Smirnoff test with statistic of 0.42 and $\tx{p-value} \ll 0.001$ on RDW distributions). We again consider the previously mentioned delay between venepuncture and measurement and the time into the year as continuous domains. As classification tasks, we selected patient sex, age bracket, and care setting (i.e. whether they are treated in primary or secondary care). The sample distributions for domains and tasks are displayed in Appendix~\ref{sec:A_EpiCov_FBC_preprocess}.

\subsection*{Proposed Disentangled Autoencoder Model}\label{sec:Dis-AE_model}
\begin{figure}[h]
\centering
\includegraphics[width=.9\linewidth]{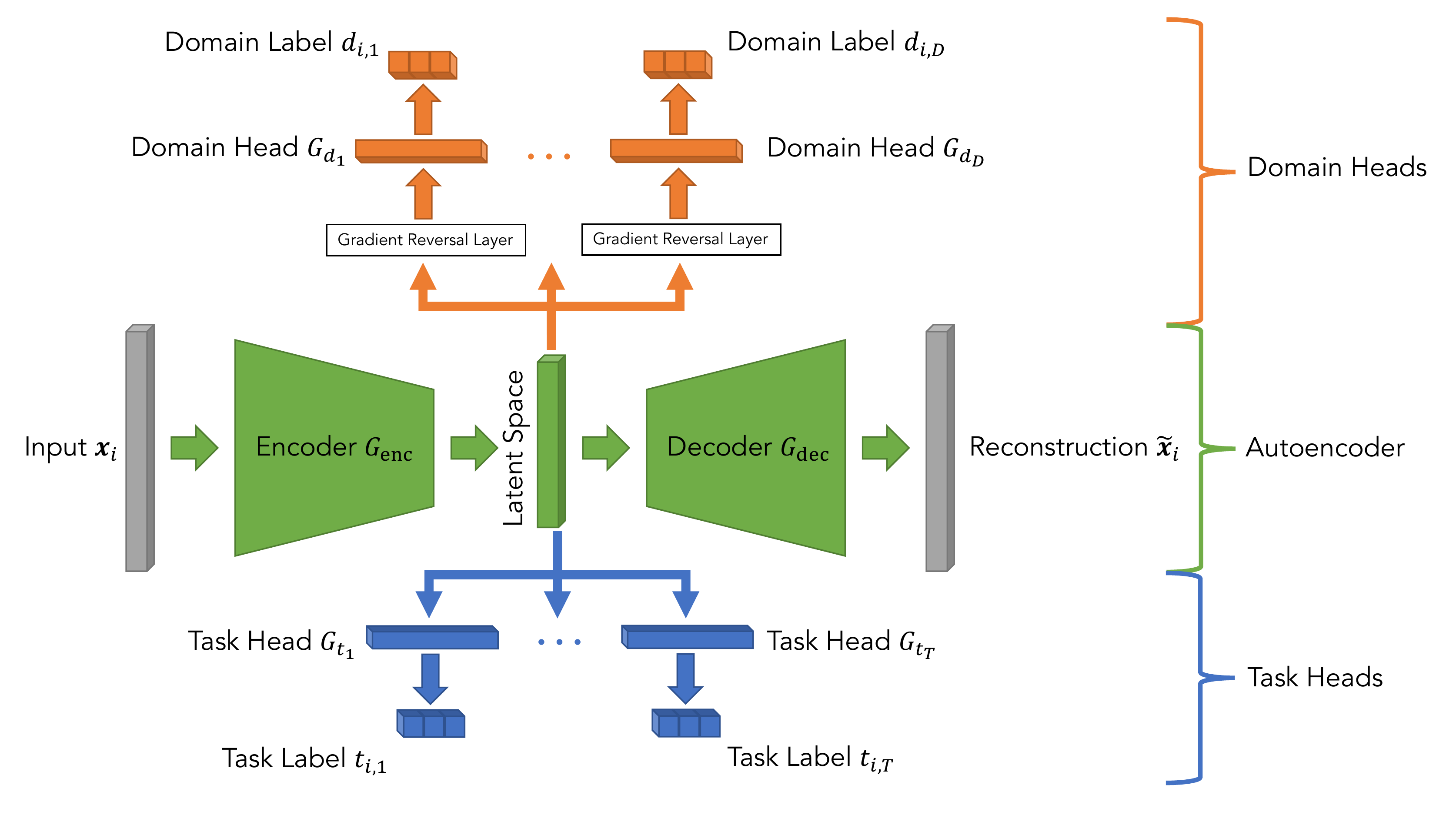}
\caption{\label{fig:DisAE_schematic}
The proposed disentangled autoencoder (Dis-AE) architecture includes an autoencoder as well as multiple task and domain heads attached to the autoencoder latent space. We train the model to maximise the performance of the task heads and decoder reconstruction while making the domain label prediction of the domain heads as difficult as possible. This adversarial training enables the model to learn a disentangled representation for multiple domains while ensuring good classification performance on multiple tasks.}
\end{figure}
\noindent
We seek to achieve domain generalisation using a domain-adversarial learning approach. In such an approach, adversarial learning is used to find domain-agnostic features from which the source domain label of the data can no longer be predicted. 
We chose this method as we found it flexible, allowing for multiple domains and tasks. This framework also allows us to introduce an autoencoder architecture, thereby ensuring that the domain-agnostic learned embeddings of the data are also meaningful.
Such embeddings cluster semantically similar inputs, making them valuable across various downstream models.
Our proposed disentangled autoencoder (Dis-AE) model is largely inspired by Ganin \& Lempitsky~\cite{pmlr-v37-ganin15}, who proposed a model for domain adaptation purposes. They trained an adversarial model where a feature extractor and a class label predictor attempt to fool a domain classifier, reminiscent of other adversarial model concepts \cite{goodfellowGenerativeAdversarialNetworks2020}. The key difference to other adversarial training schemes was the introduction of a \textit{gradient reversal layer} which enables training all model parameters in one optimisation step rather than alternating between the adversarial parties. Trivedi et al.~\cite{trivediDeepLearningModels2022} later repurposed this general idea for domain generalisation for medical images to obtain disentangled image representations which do not contain image source information.\\
We extend this concept to tabular medical data and expand its capabilities in two major ways. Firstly, we incorporate a decoder network mirroring the feature extractor, to ensure the model learns to extract meaningful representations from the data. Secondly, we consider a more realistic separation when describing a dataset's domain shift (see Figure~\ref{fig:domain_subdomain}) and introduce the option to add multiple \textit{domain classifier heads} to the latent space of the model. Each domain head trains simultaneously and attempts to predict the corresponding domain label. Crucially, we allow for the classification of categorical or ordinal instance labels and regression to a continuous value label by discretising continuous domains into bins.
We similarly propose adding multiple \textit{task classifier heads} to the latent space, representing multiple classification tasks on the data.
Figure~\ref{fig:DisAE_schematic} shows our proposed model architecture.\\
More formally, we consider our input space to be made up of $N$ samples $\mathbf{x}\in X \subset \mathbb{R}^k$, which are one-dimensional vectors of tabular data with $k$ features. We also assume that each sample $\mathbf{x}_i$ has associated task labels $\{t_{i,1}, \ldots , t_{i,T}\}$ and domain labels $\{d_{i,1}, \ldots , d_{i,D}\}$, where $T$ and $D$ are the number of tasks and domains, respectively. \textit{Tasks} are outcomes of interest (e.g. sample classification, patient illness) while \textit{domains} are covariates of the data irrelevant to the tasks (see also Figure~\ref{fig:domain_subdomain}). We aim to obtain a domain-agnostic data representation.
We assume, as is standard in the literature, that the data's source and target distribution are similar but distinct and only differ by the domain shifts \cite{zhouDomainGeneralizationSurvey2022}. At training time, the model only has access to source domain distributions. We aim to train a model with high classification performance on the task labels and low autoencoder reconstruction error when given only samples from the target distribution.\\
For the proposed encoder $G_\tx{enc}$, decoder $G_\tx{dec}$ architecture, with task heads $\{G_{t_j}\}$ and domain heads $\{G_{d_j}\}$, we consider the empirical error
\begin{align}
    E(\theta_\tx{enc}, \theta_\tx{dec}, \{\theta_t\}, \{\theta_d\}) =\; &\alpha \cdot\underbrace{\sum_{i=1}^N L_\tx{MSE} \left(G_\tx{dec} \left(G_\tx{enc} (\mathbf{x}_i;\theta_\tx{enc}); \theta_\tx{dec} \right), \mathbf{x}_i \right)}_\tx{reconstruction loss} \nonumber \\
    &+ \beta \cdot \sum_{j=1}^T \underbrace{\sum_{i=1}^N L_{t_j} \left(G_{t_j} \left(G_\tx{enc}(\mathbf{x}_i; \theta_\tx{enc}); \theta_{t_j} \right), t_{i,j} \right)}_\tx{task loss for task $j$} \nonumber \\
    &+ \sum_{j=1}^D \underbrace{\sum_{i=1}^N L_{d_j} \left(G_{d_j} \left(R_\lambda \left(G_\tx{enc}(\mathbf{x}_i;\theta_\tx{enc})\right); \theta_{d_j} \right), d_{i,j} \right)}_\tx{domain loss for domain $j$}\nonumber \\
    \equiv &\; \mathcal{L}_\tx{rec.} + \mathcal{L}_\tx{task} + \mathcal{L}_\tx{domain} \; , \label{eq:loss_function}
\end{align}
where $R_\lambda$ is the gradient reversal layer that is defined by the pseudo-function
\begin{align}
    R_\lambda(\mathbf{x}) &= \mathbf{x}\\
    \frac{\total R_\lambda}{\total x} &= -\lambda \mathbf{I} \; ,
\end{align}
where $\mathbf{I}$ is the identity matrix \cite{pmlr-v37-ganin15}. The $L_{t_j}$ and $L_{d_j}$ terms represent the loss function of the $j$-th task and domain, respectively. For the classification objectives (i.e. task and domain labels), we use cross entropy loss with a softmax activation (we have not used ordinal losses in this work, though implementation would be trivial), while for autoencoder reconstruction, we use the mean-squared error loss $L_\tx{MSE}$. The tuneable hyperparameters $\alpha$, $\beta$, and $\lambda$ control the importance of each of the three training objectives.
We aim to find a saddle point of $E$ such that
\begin{align}
    \hat{\theta}_\tx{enc} &= \argmin_{\theta_\tx{enc}} \; \mathcal{L}_\tx{rec.} + \mathcal{L}_\tx{task} - \mathcal{L}_\tx{domain} \label{eq:enc_argmin}\\
    \hat{\theta}_\tx{dec} &= \argmin_{\theta_\tx{dec}} \; \mathcal{L}_\tx{rec.} \label{eq:dec_argmin}\\
    \{\hat{\theta}_t\} &= \argmin_{\{\theta_t\}} \; \mathcal{L}_\tx{task} \label{eq:t_argmin}\\
    \{\hat{\theta}_d\} &= \argmin_{\{\theta_d\}} \; \mathcal{L}_\tx{domain} \label{eq:d_argmin} \; .
\end{align}
The inclusion of gradient reversal layers allows running gradient descent-based minimisation on $\mathcal{L}_\tx{rec.} + \mathcal{L}_\tx{task} + \mathcal{L}_\tx{domain}$ with respect to all parameters $\left( \theta_\tx{enc}, \theta_\tx{dec}, \{\theta_t\}, \{\theta_d\} \right)$ at once to find the optimal parameters \eqref{eq:enc_argmin}--\eqref{eq:d_argmin} \cite{pmlr-v37-ganin15}. 

\subsection*{Out-of-Distribution Evaluation}
Ye et al.~\cite{yeTheoreticalFrameworkOutofDistribution2021} developed a theoretical framework on what it means for an out-of-distribution (OOD) problem to be \textit{learnable} by a model. We adapt their model selection method and use their definition of model variation as a baseline for our reliability assessment.

\subsubsection*{Measuring Disentanglement}\label{sec:selection_score}
For a feature $\phi(\mathbf{x})$ in the latent space generated by $G_\tx{enc} (\mathbf{x})$ and a dataset with domain and task labels as described above, the \textit{variation} of feature $\phi$ for task $\tau$ across domain $\delta$ is given by
\begin{align}
    \mathcal{V}(\phi, \tau, \delta) &:= \max_{(y, e, e') \in \{t_{\tau}\} \times \{d_{\delta}\} \times \{d_{\delta}\}} \; \rho \left( P(\phi^{e}|y), P(\phi^{e'}|y)\right) \; ,
\end{align}
where $\phi^e$ is the latent feature $\phi$ of all data with label $e$ for domain $\delta$,
$\rho(P,Q)$ is a symmetric function with which we measure dissimilarity between distributions $P$ and $Q$, and $\{t_{\tau}\}$ and $\{d_{\delta}\}$ are the unique labels of task $\tau$ and domain $\delta$, respectively. For the purpose of model selection, we use the 1-dimensional Wasserstein-2 ($\rho_{\tx{W}_2}$) metric, implemented in the Python Optimal Transport (POT)~\cite{flamary2021pot} package,
\begin{align}
    \rho_{\tx{W}_2}(P,Q) &= \sqrt{\min_{\gamma\in\mathbb{R}^{m\times n}_+} \sum_{i,j} \gamma_{ij} (x_i - y_j)^2}\\
    & \tx{s.t.} \; \gamma\mathbf{I}=P; \gamma^\top \mathbf{I} = Q; \gamma \geq 0 \; ,
\end{align}
where $x\in P$, $y\in Q$ and $m$ and $n$ are the number of data points in $P$ and $Q$, respectively. 
We use $\rho_{\tx{W}_2}$ for this task since it monotonically increases with the distance between distributions. In contrast, other metrics such as the total variation distance saturate as soon as the distributions no longer overlap. Due to the use of this metric in hyperparameter tuning, a continuous dependence on the distance between distributions is desirable.
Overall, $\mathcal{V}_\rho(\phi, t, d)$ describes the variation of a feature in the latent space across a domain. Hence, $\mathcal{V}$ should be small in a disentangled representation. Due to various scales within features, in practice, we first normalise the feature distribution to zero-mean and unit standard deviation, and then calculate $\rho \left( P(\phi^{e}|y), P(\phi^{e'}|y)\right)$.\\
Since Ye et al.~\cite{yeTheoreticalFrameworkOutofDistribution2021} only consider one task and domain, we define feature variation across all tasks and domains as
\begin{align}
    V(\phi) &:= \max_{\substack{\tau \in \{1, \ldots, T\} \\ \delta \in \{1, \ldots, D\}}} \mathcal{V}(\phi, \tau, \delta) \; .
\end{align}
To get a measure of the variation over the full latent space, we use the \textit{model variation} of the encoder:
\begin{align}\label{eq:model_variation}
    V^{\sup}(G_\tx{enc}) &:= \sup_{u \in \mathcal{S}^{m-1}} V(u^\top G_\tx{enc}(\mathbf{X})) \; ,
\end{align}
where $G_\tx{enc}(\mathbf{X})$ is the list of latent space representations of all $\mathbf{x} \in \mathbf{X}$, $m$ is the latent space dimension, and $\mathcal{S}^{m-1}$ is the unit $(m-1)$-sphere, and therefore $u^\top G_\tx{enc}(\mathbf{X})$ is a vector. We can use $V^{\sup}(G_\tx{enc})$ as an intuitive measure of how much the latent space varies across source domains. This allows us to measure how successfully domains have been disentangled.

\subsubsection*{Model Selection Score}
To select the best-performing model hyperparameters, we required a combined measure of how well a model performs on all three training objectives (reconstruction, task classification, domain disentanglement).
Ye et al.~\cite{yeTheoreticalFrameworkOutofDistribution2021} suggest a score comprised of a model's classification performance and its variation as a metric for model selection. Since we additionally aim for good reconstruction performance, our proposed \textbf{Selection Score} for a model $f$ instead consists of three terms:
\begin{align}\label{eq:selection_score}
    \tx{Score}(f) := \underbrace{\frac{1}{T} \sum_{\tau \in \{1, \ldots, T\}} \tx{Acc}_{\tau,f}}_\tx{mean task accuracy} - \underbrace{\frac{V^{\sup}(G_{\tx{enc},f})}{V^{\sup}(\mathbf{I})}}_\tx{rel. model variation} - \underbrace{\sqrt{\frac{\tx{MSE}(\mathbf{X}, \tilde{\mathbf{X}})}{\mathbb{E}(\mathbf{X}^2)}}}_\tx{rel. rec. error} \; ,
\end{align}
where $\tx{MSE}(\mathbf{X}, \tilde{\mathbf{X}})$ is the mean-squared error between input data $\mathbf{X}$ and its reconstruction $\tilde{\mathbf{X}}$ by the model and $\mathbb{E}(\mathbf{X}^2)$ is the mean of all squared samples $\lVert \mathbf{x} \rVert_2^2$. $V^{\sup}(\mathbf{I})$ gives the domain variation of the data without any embedding transformation.\\
Since the second and third terms in Eq.~\eqref{eq:selection_score} are unbounded (depending on choice of $\rho$), $\tx{Score}(f) \in (-\infty, 1]$, where a score close to 1 is desirable.

\subsubsection*{Reliability Assessment}\label{sec:method_reliability_assessment}
The variation of a model, as defined in Eq.~\eqref{eq:model_variation}, can be used to quantify the dissimilarity between domains for any dataset, whether it is in full-dimensional space or a lower-dimensional embedding space.
A successful domain generalisation model should maintain similar model variation on source and target data.
For reliability assessment experiments, we used the Jensen-Shannon divergence
\begin{align}\label{eq:jensen-shannon}
    \rho_\tx{JSD}(P,Q) &= \frac{1}{2} D(P,M) + \frac{1}{2} D(Q,M) \; ,
\end{align}
instead of the Wasserstein distance in the calculation of $V^{\sup}$. Here, $D(P,Q)$ is the Kullback-Leibler divergence~\cite{CoverT.M.2006Eoit} and $M=\frac{1}{2}(P+Q)$. This was chosen as
$\rho_\tx{JSD}(P, Q)$ is the reduction in uncertainty of a random variable $X$, given knowledge of whether it originated from $P$ or $Q$ (mutual information \cite{CoverT.M.2006Eoit}). Using the Jensen-Shannon divergence in the calculation of $V^{\sup}$ is thus a direct measurement of how much
domain information is contained within our data representation. For $V^{\sup}$ close to 0, there is no domain shift; for $V^{\sup}$ close to 1, the distributions are completely separated.

\subsection*{Architecture Selection for Dis-AE}\label{sec:Dis-AE_performance}
The Dis-AE architecture may be viewed as a “vanilla” autoencoder (AE) with task and domain heads. Therefore, to determine an optimal Dis-AE architecture, we begin by identifying the best AE structure.
In all experiments, we perform hyperparameter tuning on a vanilla AE with fully-connected layers to optimise the network's depth and width. We then optimise the hyperparameters of the Dis-AE model with respect to the selection score from Eq.~\eqref{eq:selection_score}, while keeping the network architecture (i.e., depth and width) the same as the vanilla AE.
The AE consists of fully-connected linear layers, ReLU activations, and a learning rate scheduler that reduces the learning rate when the validation loss reaches a plateau. We perform hyperparameter tuning of the Dis-AE model for the $\alpha$, $\beta$, and $\lambda$ parameters (see Eq.~\eqref{eq:loss_function}) and an L2-regularisation parameter. We perform hyperparameter tuning using the Weights and Biases~\cite{wandb} framework and the Cambridge Service for Data-Driven Discovery (CSD3) high-performance computing platform, exhaustively searching among parameter combinations.\\
Throughout this work, all reported experimental results were obtained as the average of a 5-fold cross-validation experiment. Additionally, we trained the models 5 times per fold and selected the best performing models for each fold. This was done to mitigate differences due to weight initialisation.

\clearpage
\section{Results}

\subsection*{Performance of the Proposed Model}
\begin{table}[h]
\begin{center}
\begin{tabularx}{\textwidth}{c|c|ccc|c}
\toprule
        Data & Model & Acc. Term & Var. Term & Rec. Term & Score\\
        \midrule
        Data A & Vanilla AE & 0.97 & 1.011 & 0.05 & -0.09\\
         & Dis-AE & 0.98 & 0.003 & 0.08 & \textbf{0.89}\\
         \midrule
        Data B & Vanilla AE & 0.96 & 0.693 & 0.06 & 0.21\\
         & Dis-AE & 0.98 & 0.011 & 0.17 & \textbf{0.80}\\
        \midrule
        Data C & Vanilla AE & \{0.77, 0.73, 0.71\} & 0.633 & 0.24 & -0.14\\
        & Dis-AE & \{0.81, 0.77, 0.73\} & 0.074 & 0.37 & \textbf{0.33}\\
        \midrule
        INTERVAL & Vanilla AE & \{0.85, 0.28, 0.51\} & 1.993 & 0.16 & -1.61\\
         & Dis-AE & \{0.85, 0.28, 0.50\} & 0.219 & 0.21 & \textbf{0.11}\\
        \midrule
        CUH & Vanilla AE & \{0.75, 0.24, 0.77\} & 1.089 & 0.02 & -0.52\\
         & Dis-AE & \{0.75, 0.24, 0.74\} & 0.631 & 0.14 & \textbf{-0.19}\\
\bottomrule
\end{tabularx}
\caption{Task classification accuracy, relative model variation, and relative reconstruction error, combined to give selection score (Eq.~\eqref{eq:selection_score}). The results were obtained from experiments on various datasets using a vanilla AE and the Dis-AE model.
For the synthetic datasets, instances 1\&2 of domain 1 were selected as source data (see Section~\ref{sec:datasets}). For multi-task problems, the accuracy per task is displayed.
The tasks for the INTERVAL dataset are subject \{sex, age bracket, and BMI bracket\}.
For the CUH dataset, the two analysers with the highest variation HL-FBC data were selected as source data. The tasks for the CUH dataset are patient \{sex, age bracket, and care setting\}. During training, we used weighted random sampling to create batches with balanced care setting numbers.
}\label{tab:selection_scores}
\end{center}
\end{table}
\noindent
The selection score given in Eq.~\eqref{eq:selection_score} (and its constituent terms) give an illustrative metric of the model's performance. Table~\ref{tab:selection_scores} displays these terms for both the vanilla AE and Dis-AE models for all datasets. Further experimental results, including task classification accuracies for the various datasets, are included in Appendix~\ref{sec:A_Dis-AE_performance}, where we see similar behaviour.
An illustration of generalisation performance is given by Table~\ref{tab:FBC_variations}, where domain generalisation performance on INTERVAL (source) and COMPARE (target) data is displayed using the model variation (Eq.~\ref{eq:model_variation}).
\begin{table}[h]
\begin{center}
\begin{tabularx}{0.78\textwidth}{l|c|c|c}
\toprule
     & & \multicolumn{2}{c}{\textbf{Variation $V^{\sup}$}} \\
    Representation & Domain & Within source & Source and target \\
     & & (INTERVAL) & (INT $\cup$ COM)\\
     \midrule
    & Analyser & 0.50 & 0.89\\
     (A) Raw data & Sample age & 0.90 & 0.86\\
     & Time of year & 0.77 & 0.91\\
     \hline
     & Analyser & 0.46 & 0.67\\
     (B) Vanilla AE & Sample age & 0.84 & 0.83\\
     & Time of year & \textbf{0.62} & 0.85\\
     \hline
     & Analyser & \textbf{0.24} & \textbf{0.31}\\
     (C) Dis-AE & Sample age & \textbf{0.59} & \textbf{0.61}\\
     & Time of year & 0.66 & \textbf{0.74}\\
    \bottomrule
\end{tabularx}
\caption{Variation of rich FBC data from the INTERVAL and COMPARE studies in different representations. The three representations are: (A) the unmodified data, (B) the latent space of a vanilla AE and (C) the latent space of the Dis-AE model. The variation $V^{\sup}$, calculated using $\rho_\tx{JSD}$, is shown per domain to show each covariate's contribution to the data's domain shift.}
\label{tab:FBC_variations}
\end{center}
\end{table}\\

\begin{figure}
     \centering
     \begin{subfigure}[b]{0.48\textwidth}
         \centering
         \includegraphics[width=\textwidth]{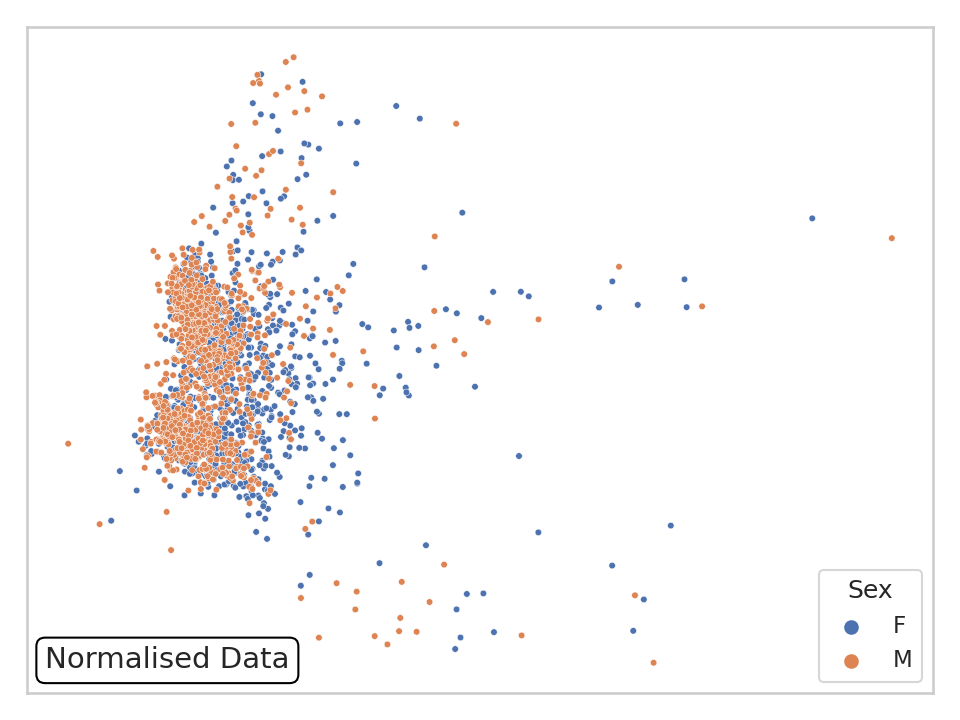}
         \caption{}
     \end{subfigure}
     \hfill
     \begin{subfigure}[b]{0.48\textwidth}
         \centering
         \includegraphics[width=\textwidth]{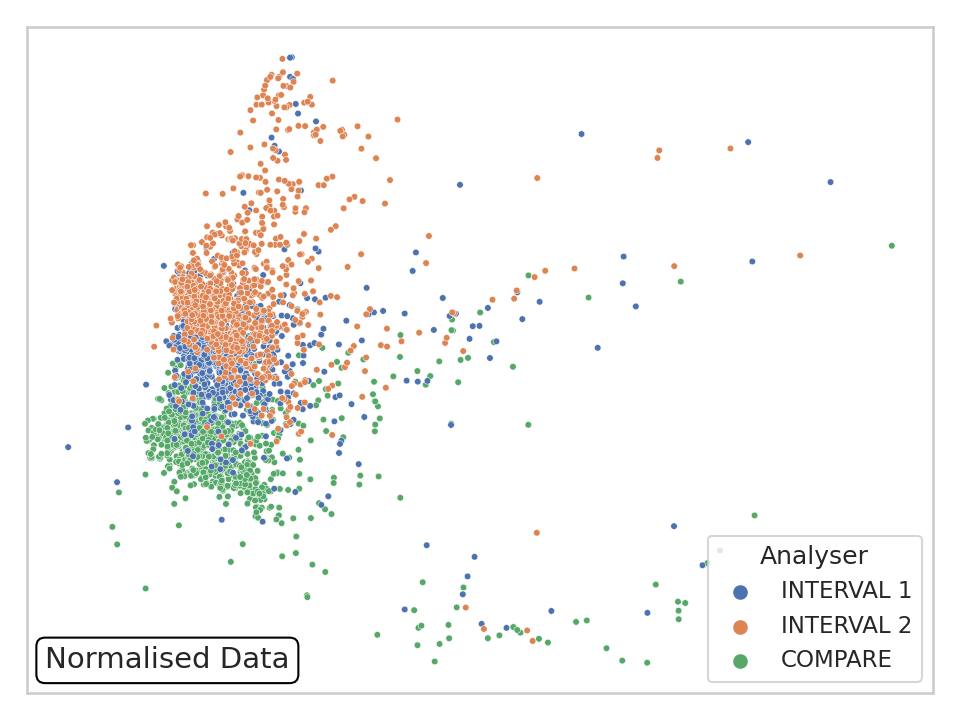}
         \caption{}
     \end{subfigure}\\
     \begin{subfigure}[b]{0.48\textwidth}
         \centering
         \includegraphics[width=\textwidth]{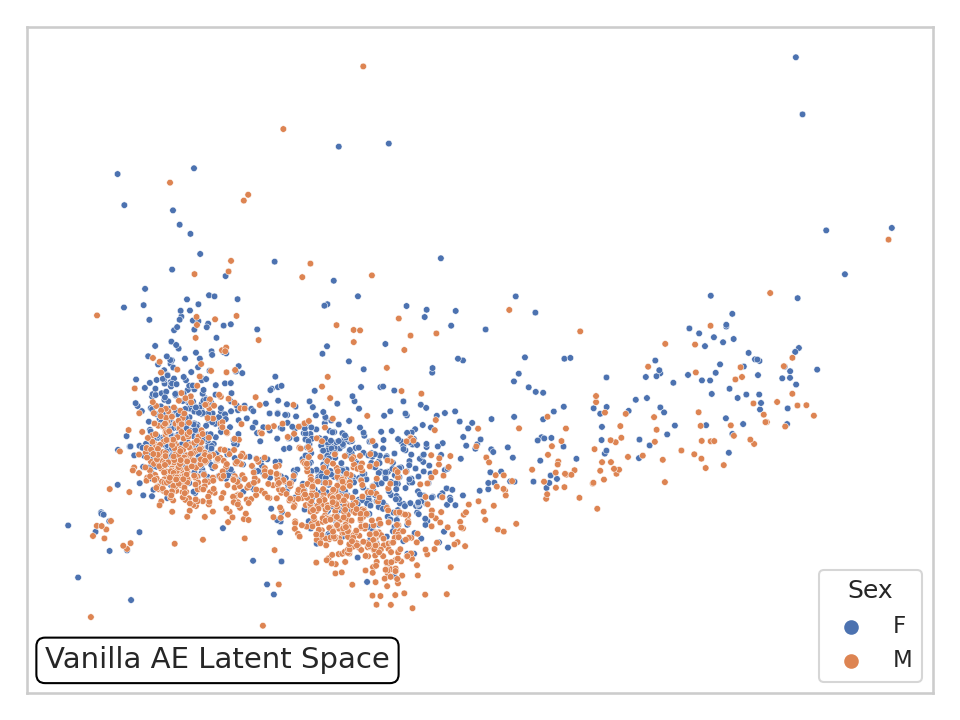}
         \caption{}
     \end{subfigure}
     \hfill
     \begin{subfigure}[b]{0.48\textwidth}
         \centering
         \includegraphics[width=\textwidth]{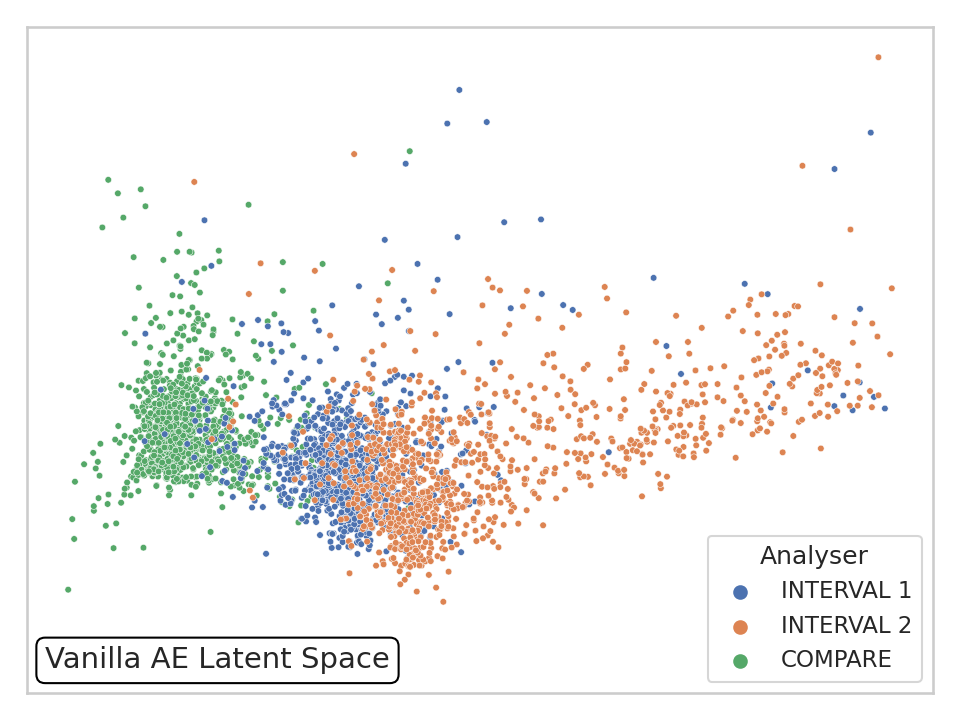}
         \caption{}
     \end{subfigure}\\
     \begin{subfigure}[b]{0.48\textwidth}
         \centering
         \includegraphics[width=\textwidth]{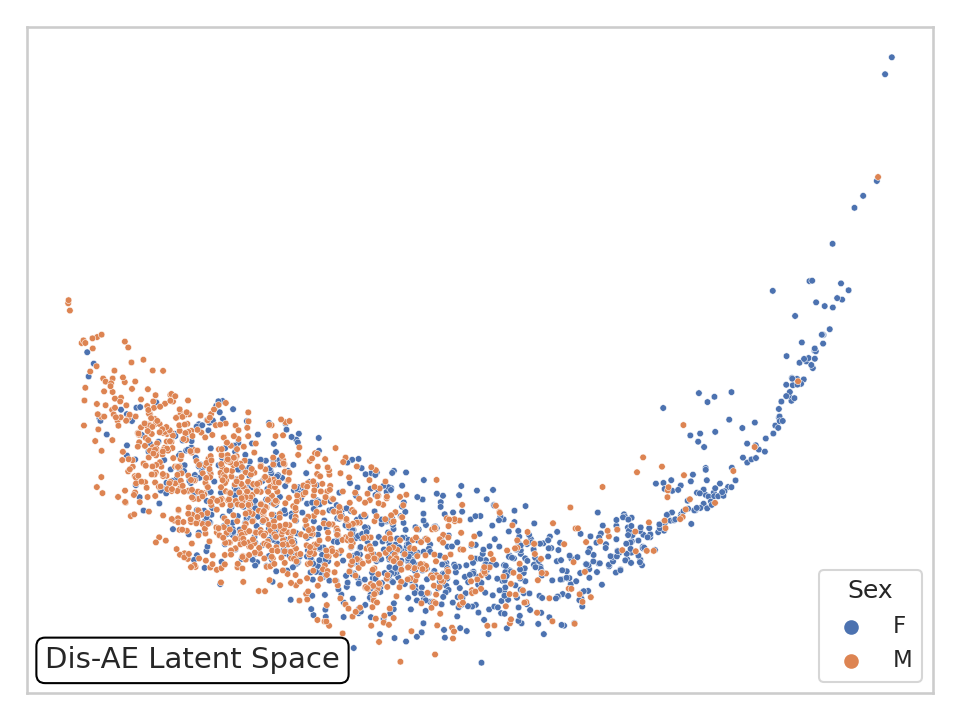}
         \caption{}
     \end{subfigure}
     \hfill
     \begin{subfigure}[b]{0.48\textwidth}
         \centering
         \includegraphics[width=\textwidth]{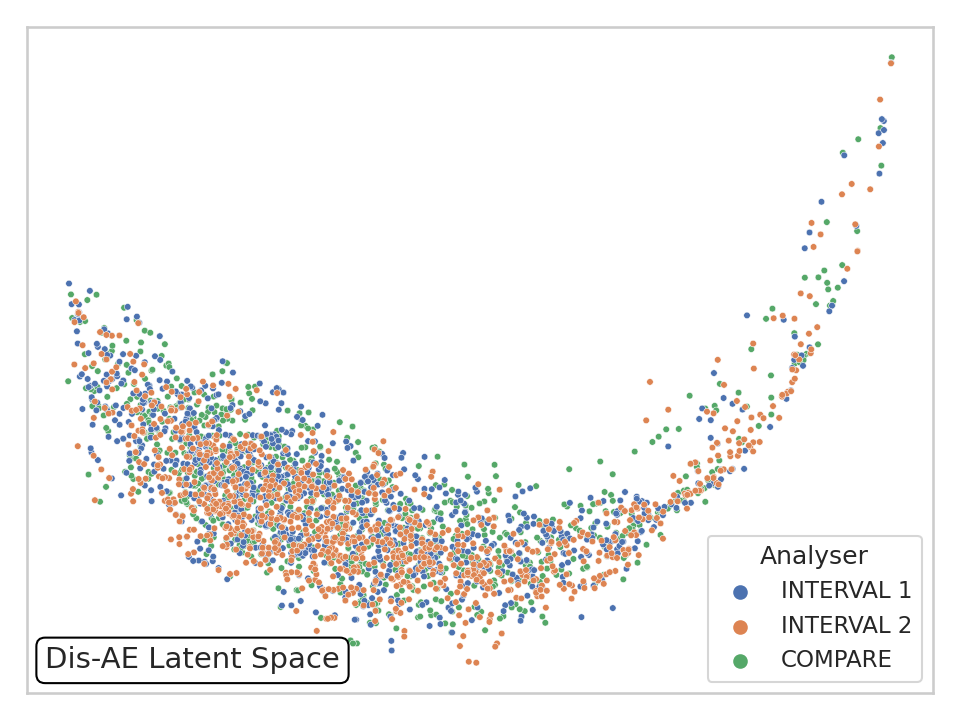}
         \caption{}
     \end{subfigure}
        \caption{Rich FBC donor data PHATE~\cite{moonVisualizingStructureTransitions2019} visualisation of normalised raw data (row 1) and latent spaces of a vanilla AE (row 2) and the Dis-AE model (row 3). The images in the left column are coloured by subject sex (example task), while those in the right column are coloured by haematology analyser (example domain). The machine learning models were only trained on data from INTERVAL.}
        \label{fig:AEvDis-AE_FBC_PHATE}
\end{figure}

Figure~\ref{fig:AEvDis-AE_FBC_PHATE} displays \textit{Potential of Heat diffusion for Affinity-based Transition Embedding} (PHATE)~\cite{moonVisualizingStructureTransitions2019} visualisations of the three representations of the rich FBC data of the INTERVAL and COMPARE studies. We trained the Dis-AE model on three tasks: subject sex, age, and BMI; and three domains: haematology analyser, the delay between venepuncture and analysis, and time of year.

\clearpage
\subsection*{Robustness Evaluation}\label{sec:Dis-AE_robustness}
\begin{figure}[h]
\centering
\includegraphics[width=\linewidth]{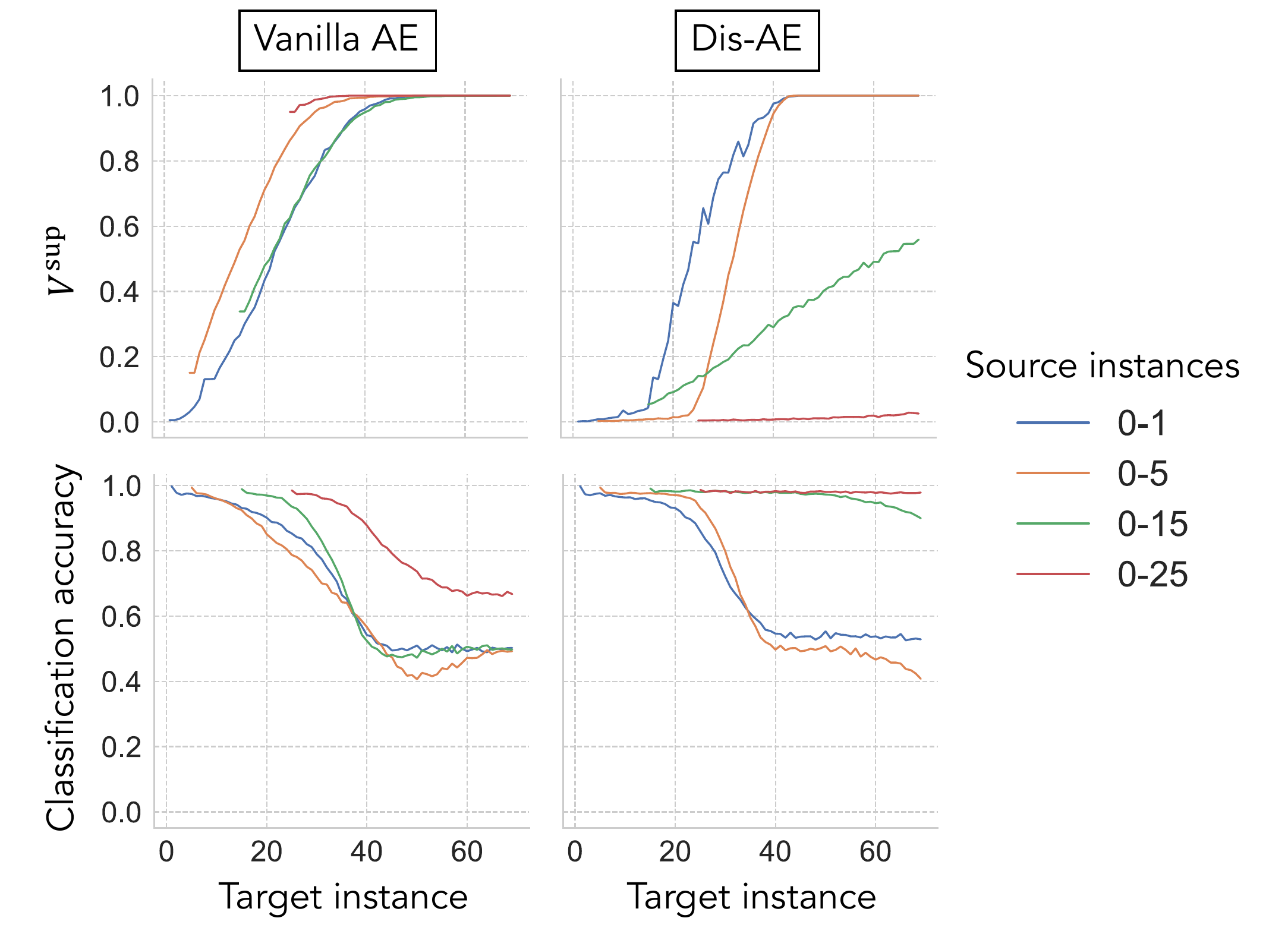}
\caption{Results of evaluating vanilla AE and Dis-AE model performance for a dataset with many instances with affine domain shift. The diversity and size of the source dataset were varied for different runs to investigate the effect on target data generalisation. The quantities shown are $V^{\sup}$ using $\rho_\tx{JSD}$ (Eq.~\eqref{eq:jensen-shannon}) and the task accuracy.
        }
        \label{fig:manymachines_results}
\end{figure}

We want to understand the extent to which the target data distribution can differ from the source distributions while preserving model robustness.
To this end, we trained a vanilla AE and the Dis-AE model multiple times, varying the number of instances included in the source data to evaluate the robustness on the target data, using the Many Affines dataset. Figure~\ref{fig:manymachines_results} shows both models' variation $V^{\sup}$ on increasingly distant target data and classification accuracy on the synthetic task.
\\
As mentioned in Section~\ref{sec:method_reliability_assessment}, we used the Jensen-Shannon divergence to calculate $V^{\sup}$ to measure mutual information between the latent space representations of the data and the source and target distributions. From the right column in Figure~\ref{fig:manymachines_results}, it is visible that training the Dis-AE model on more diverse source data results in a slower decay in task and generalisation performance. A vanilla AE does not reflect this behaviour (left column in Figure~\ref{fig:manymachines_results}).

\section{Discussion}
Our goal was to develop an approach to obtain lower-dimensional data representations that preserve informative features for outcomes of interest while eliminating influences on the data known to be irrelevant covariates. This approach could be used in a medical setting, for example, to retain clinically relevant features within the data representation while ignoring irrelevant ones, enabling training of classifiers which generalise across multiple domains. We proposed the disentangled autoencoder (Dis-AE) model as this approach and used the proposed model selection score from Eq.~\eqref{eq:selection_score} to illustrate its performance. Table~\ref{tab:selection_scores} displays the results, with the accuracy term representing the model's ability to maintain informative features within its latent space. The task performance is comparable to that of a vanilla AE. However, as supported by the results in the model variation term, Dis-AE can maintain high task performance while drastically reducing domain shift within its representation space. This improvement over a classical embedding model (vanilla AE) and the raw data is also evident from the PHATE visualisations in Figure~\ref{fig:AEvDis-AE_FBC_PHATE}. Additionally, Table~\ref{tab:FBC_variations} demonstrates Dis-AE's ability to maintain low model variation in the latent space on unseen data, even when the variation increases on the raw data.\\
We investigated the Dis-AE model's OOD robustness by measuring its performance on increasingly distant OOD target data. As seen in Figure~\ref{fig:manymachines_results}, Dis-AE maintains high task and generalisation performance even on highly OOD data, given enough diversity in the source data. This behaviour strongly suggests that using data from multiple sources can significantly increase model robustness in a real-world deployment setting.\\
Dis-AE stands out from current domain generalisation approaches as it, for the first time, considers multiple domains and tasks within its design and, as such, can separate multiple independent domain shift influences on the data. Dis-AE is scalable to many domains and tasks and can be easily modified by adding or removing classifier heads from its latent space. Including the gradient reversal layer ensures fast convergence, enabling single-step updates to all model parameters. The Dis-AE framework can be adapted to other model architectures by modifying the autoencoder part of the model to include, e.g. convolutional layers or skip-connections. Dis-AE will open up future research on data where domain shift is expected to be caused by many domains and where the exact nature of the shift is unknown. In particular, it allows for joint data embeddings where the data originated from multiple sources.\\
The proposed model has several limitations. As seen in Table~\ref{tab:selection_scores}, Dis-AE's reconstruction quality is lower than that of a vanilla AE due to conflicting objectives: removing domain information from the latent space while preserving it in the reconstruction. To overcome this, future work could involve providing the decoder with explicit domain labels, such that the decoder can rely on this additional information alongside the latent space to produce accurate reconstructions of the input data. Additionally, Dis-AE training proved somewhat unstable, converging to unsatisfactory results around 20\% of the time. Adversarial training schemes, like the one used, are notoriously unstable \cite{meschederNumericsGANs2017, arjovskyPrincipledMethodsTraining2017}. Improving convergence reliability will require further research. Fortunately, in our experiments, model training consistently completed within 10 minutes on a modern laptop, minimising concerns about re-training.

\section{Conclusion}

In this paper, we propose a novel model architecture for domain generalisation, namely a disentangled autoencoder (Dis-AE). Unlike previous work, we propose a new hierarchy for domain shift problems where we consider domains as distinct covariates which cause a shift in the data. Domains can either be characterised by continuous values (e.g. time) or consist of categorical instances (e.g. measurement device). The Dis-AE framework includes a gradient reversal layer for fast convergence and is easily adaptable to other model architectures. We compare the performance of Dis-AE to a “vanilla” autoencoder on multiple synthetic datasets and real-life clinical data. The results of the experiments demonstrate the significantly increased generalisation performance of our model and provide evidence that the Dis-AE design encourages increased model robustness when trained on more diverse training data.\\

Dis-AE provides a scalable solution for future research on large collaborative medical datasets suffering from domain shift on many domains. This may be particularly useful in multi-centre studies and trials using repeated measurements.

\subsection*{Code availability}
The code is publicly available at \textbf{[inserted on acceptance]}.

\subsection*{Availability of data and materials}
The synthetic datasets are available in the above GitLab repository. The INTERVAL and COMPARE datasets are both available from the respective study teams on reasonable request. The Epicov dataset is a private dataset held by Cambridge University Hospitals NHS Trust and is not available publicly.

\clearpage

\bibliography{references-bibtex}%

\begin{appendices}

\clearpage
\section{Synthetic dataset generation}\label{sec:A_DatasetsGeneration}
\begin{figure}[h]
\centering
\includegraphics[width=\linewidth]{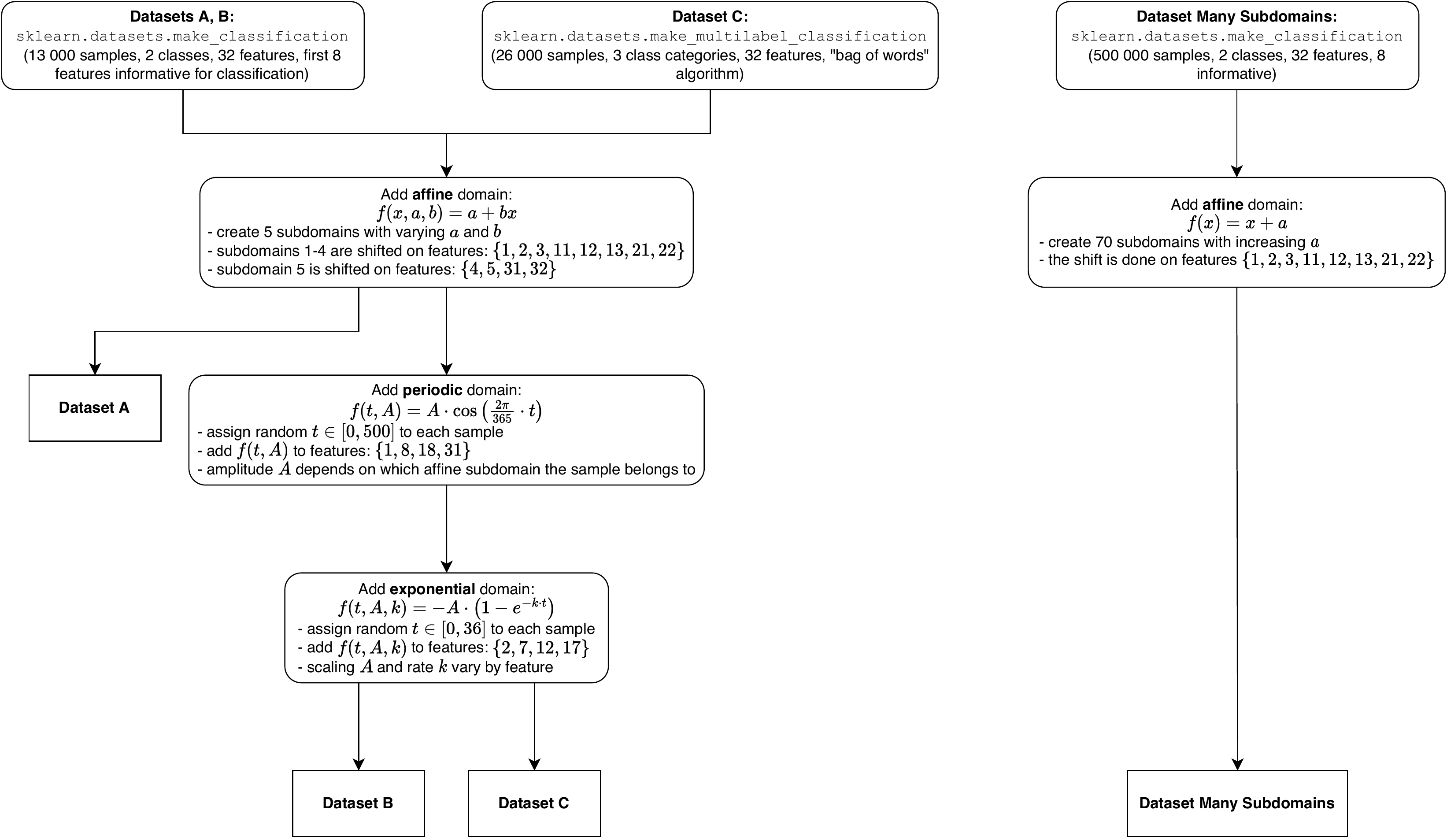}
\caption{\label{fig:synthetic_datasets_flowchart}
Overview of the method for generating the synthetic datasets. Each dataset has at least one artificial classification task and domain shift. The added domains are based on simple transformations of some data features.}
\end{figure}

\clearpage
\section{Haematological parameters}\label{sec:A_FBC}
The FBC parameters used in this work were taken directly from the corresponding datasets. Tables \ref{tab:FBC_parameters_PLT}, \ref{tab:FBC_parameters_RBC} and \ref{tab:FBC_parameters_WBC} list and describe the used HL-FBC and R-FBC parameters. The parameters correspond to standard FBC measurements as well as additional research and clustering parameters recorded by the Sysmex XN-2000 haematology analysers of the INTERVAL and COMPARE trials (see also Section~\ref{sec:datasets}).
\begin{table}[h]
\begin{center}
\caption{List of used platelet features. (HL-FBC features in bold)}\label{tab:FBC_parameters_PLT}%
\begin{tabularx}{\textwidth}{ l X l }
\toprule
        Sysmex Abbreviation & Long Name & Unit \\\midrule
         \textbf{PLT} & \textbf{Platelet count} & \textbf{per nL} \\
         PDW & Platelet distribution width & fL \\
         \textbf{MPV} & \textbf{Mean platelet volume} & \textbf{fL} \\
         P-LCR & Platelet - large cell ratio & \% \\
         \textbf{PCT} & \textbf{Plateletcrit} & \textbf{\%} \\
         IPF & Immature platelet fraction & - \\
         PLT-I & PLT count calculated from the PLT particle size distribution & per nL \\
         PLT-O & PLT count calculated from the RET channel & per nL \\
         PLT-F & PLT count calculated from the PLT-F channel & per nL \\
         H-IPF & High fluorescence immature platelet fraction & - \\
         IPF\# & Immature platelet count & per nL \\
        \botrule
\end{tabularx}
\end{center}
\end{table}
\begin{table}[h]
\begin{center}
\caption{List of used red cell features. (HL-FBC features in bold)}\label{tab:FBC_parameters_RBC}
\begin{tabularx}{\textwidth}{ l X l }
\toprule
        Sysmex Abbreviation & Long Name & Unit \\\midrule
         \textbf{RBC} & \textbf{Red blood cell count} & \textbf{per pL} \\
         \textbf{HGB} & \textbf{Haemoglobin concentration} & \textbf{g/dL} \\
         \textbf{HCT} &\textbf{Haematocrit} & \textbf{\%} \\
         \textbf{MCV} & \textbf{Mean corpuscular volume} &\textbf{fL} \\
         \textbf{MCH} & \textbf{Mean corpuscular haemoglobin} & \textbf{pg} \\
         \textbf{MCHC} & \textbf{Mean corpuscular haemoglobin concentration} &\textbf{g/dL} \\
         \textbf{RDW-SD} & \textbf{Red cell distribution width (standard deviation)} & \textbf{fL} \\
         RDW-CV & Red cell distribution width (coefficient of variation) & \% \\
         RET\% & Reticulocyte percent & \% \\
         RET\# & Reticulocyte count & per pL \\
         IRF & Immature reticulocyte fraction & \% \\
         LFR & Low fluorescence ratio & \% \\
         MFR & Medium fluorescence ratio & \% \\
         HFR & High fluorescence ratio & \% \\
         RET-He & Reticulocyte haemoglobin equivalent & pg \\
         MicroR & Micro RBC ratio & \% \\
         MacroR & Macro RBC ratio & \% \\
         RBC-O & RBC count calculated from the RET channel & per pL \\
         RBC-He & Red blood cell haemoglobin equivalent & pg \\
         Delta-He & Delta of the haemoglobin equivalents & pg \\
         RET-Y & Forward-scatter in RET area of RET scattergram & ch \\
         RET-RBC-Y & Forward-scatter in RBC area of RET scattergram & ch \\
         IRF-Y & Forward-scatter in IRF area of RET scattergram & ch \\
         Hypo-He & Hypo-haemoglobinised percentage of red cells & \% \\
         Hyper-He & Hyper-haemoglobinised percentage of red cells & \% \\
         RPI & Reticulocyte Production Index & - \\
         RET-UPP & Count in the UPP area of the RET scattergram & - \\
         RET-TNC & Count in the TNC area of the RET scattergram & - \\
        \botrule
\end{tabularx}
\end{center}
\end{table}
\begin{table}[h]
\begin{center}
\caption{List of used white cell features. (HL-FBC features in bold)}\label{tab:FBC_parameters_WBC}
\begin{tabularx}{\textwidth}{ l X l }
\toprule
        Sysmex Abbreviation & Long Name & Unit \\\midrule
         \textbf{WBC} & \textbf{White blood cell count} &\textbf{per nL} \\ 
         \textbf{NEUT\#} & \textbf{Neutrophil count} & \textbf{per nL} \\
         \textbf{LYMPH\#} & \textbf{Lymphocyte count} & \textbf{per nL} \\
         \textbf{MONO\#} & \textbf{Monocyte count} & \textbf{per nL} \\
        \textbf{EO\#} & \textbf{Eosinophil count} & \textbf{per nL} \\
         \textbf{BASO\#} & \textbf{Basophil count} & \textbf{per nL} \\
         \textbf{NEUT\%} & \textbf{Neutrophil percent} & \textbf{\%} \\
         \textbf{LYMPH\%} & \textbf{Lymphocyte percen}t & \textbf{\%} \\
         \textbf{MONO\%} & \textbf{Monocyte percent} &\textbf{\%} \\
         \textbf{EO\%} &\textbf{Eosinophil percent} & \textbf{\%} \\
         \textbf{BASO\%} & \textbf{Basophil percent} & \textbf{\%} \\
         IG\# & Immature granulocyte count & per nL \\
         IG\% & Immature granulocyte percent & \% \\
         TNC & Total nuclear cell count & per nL \\
         WBC-N & WBC count calculated from WNR channel & per nL \\
         TNC-N & Total nuclear cell count calculated from WNR channel & per nL \\
         BA-N\# & Basophil count calculated from the WNR channel & per nL \\
         BA-N\% & Basophil percent calculated from the WNR channel & \% \\
         WBC-D & WBC count calculated from WDF channel & per nL \\
         TNC-D & Total nuclear cell count calculated from WDF channel & per nL \\
         NEUT\#\& & Subtraction of IG\# from NEUT\# & per nL \\
         NEUT\%\& & Ratio of NEUT\#\& to WBC & \% \\
         LYMP\#\& & Subtraction of the upper LYMPH AWS from LYMPH\# & per nL \\
         LYMP\%\& & Ratio of LYMP\#\& to WBC & \%\\
         BA-D\# & Basophil count calculated from the WDF channel & per nL \\
         BA-D\% & Basophil percent calculated from the WDF channel & \% \\
         NE-SSC & Lateral scattered light intensity of the NEUT AWS & ch \\
         NE-SFL & Fluorescent light intensity of the NEUT AWS & ch \\
         NE-FSC & Forward-scattered light intensity of the NEUT AWS & ch \\
         LY-X & Lateral scattered light intensity of the LYMPH AWS & ch \\
         LY-Y & Fluorescent light intensity of the LYMPH AWS & ch \\
         LY-Z & Forward-scattered light intensity of the LYMPH AWS & ch \\
         MO-X & Lateral scattered light intensity of the MONO AWS & ch \\
         MO-Y & Fluorescent light intensity of the MONO AWS & ch \\
         MO-Z & Forward-scattered light intensity of the MONO AWS & ch \\
         NE-WX & Lateral scattered light distribution width of the NEUT AWS & \textperthousand\\
         NE-WY & Fluorescent light distribution width of the NEUT AWS & \textperthousand\\
         NE-WZ & Forward-scattered light distribution width of the NEUT AWS & \textperthousand\\
         LY-WX & Lateral scattered light distribution width of the LYMPH AWS & \textperthousand\\
         LY-WY & Fluorescent light distribution width of the LYMPH AWS & \textperthousand\\
         LY-WZ & Forward-scattered light distribution width of the LYMPH AWS & \textperthousand\\
         MO-WX & Lateral scattered light distribution width of the MONO AWS & \textperthousand\\
         MO-WY & Fluorescent light distribution width of the MONO AWS & \textperthousand\\
         MO-WZ & Forward-scattered light distribution width of the MONO AWS & \textperthousand\\
         \botrule
\end{tabularx}
\footnotetext{AWS: area on the white cell differential fluorescence scattergram}
\end{center}
\end{table}

\clearpage
\section{Preparation of FBC data from the INTERVAL and COMPARE trials}\label{sec:A_FBC_preprocess}
\begin{figure}[h]
\centering
\includegraphics[width=\linewidth]{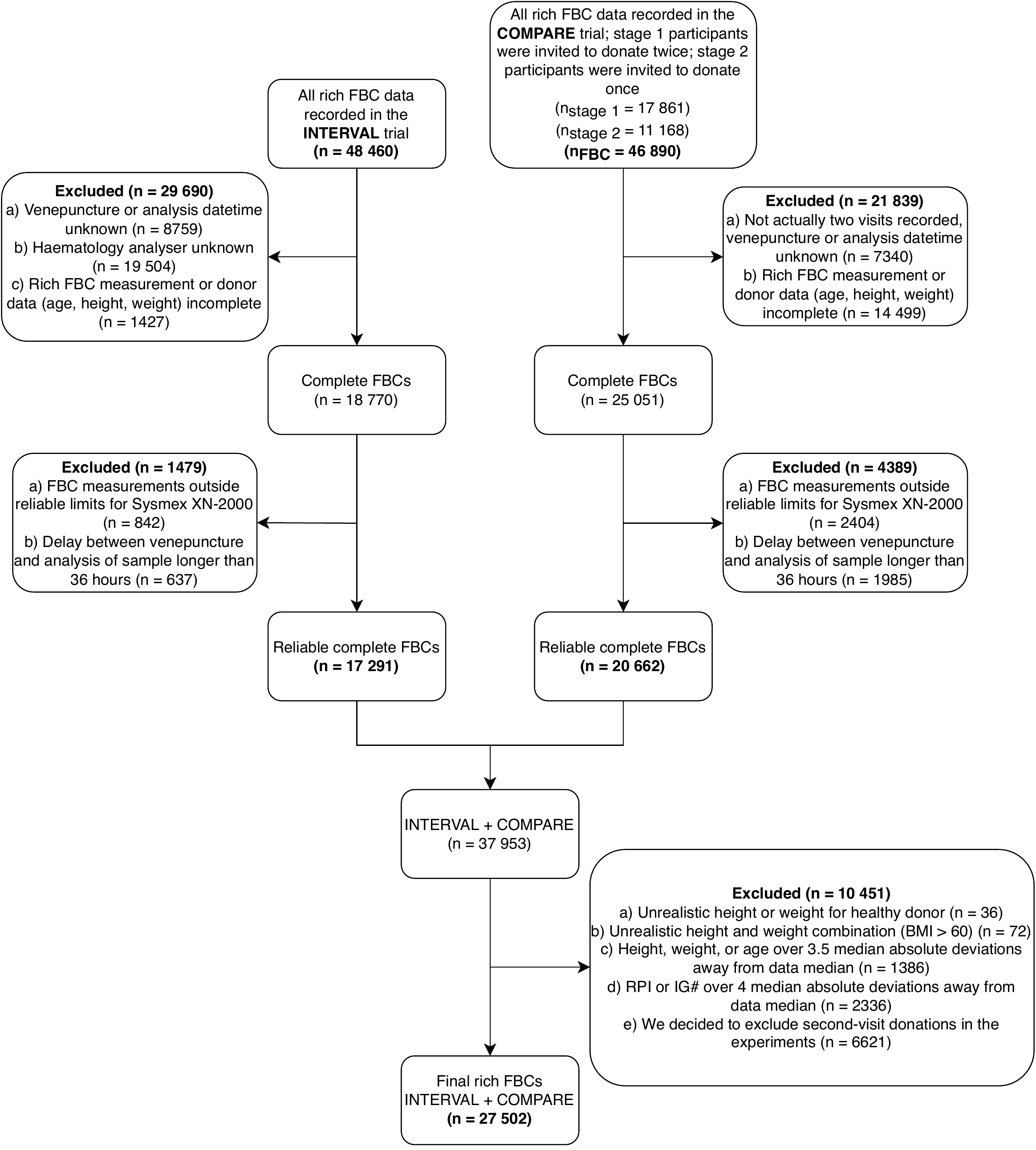}
\caption{\label{fig:INTERVAL_COMPARE_flowchart}
Flowchart of the pre-filtering of the available rich FBC data from the INTERVAL and COMPARE trials for the experiments in this paper. We used Vis \& Huisman's~\cite{visVerificationQualityControl2016} suggested reliability limits for FBC quality control as exclusion criteria.
Similarly, we only considered samples with a subject height within \qtyrange{1}{2.2}{\meter}, subject weight within \qtyrange{45}{200}{\kilogram}, and subject BMI less than 60. We only used first-visit donations per donor.
}
\end{figure}

\begin{figure}[h]
\centering
\includegraphics[width=\linewidth]{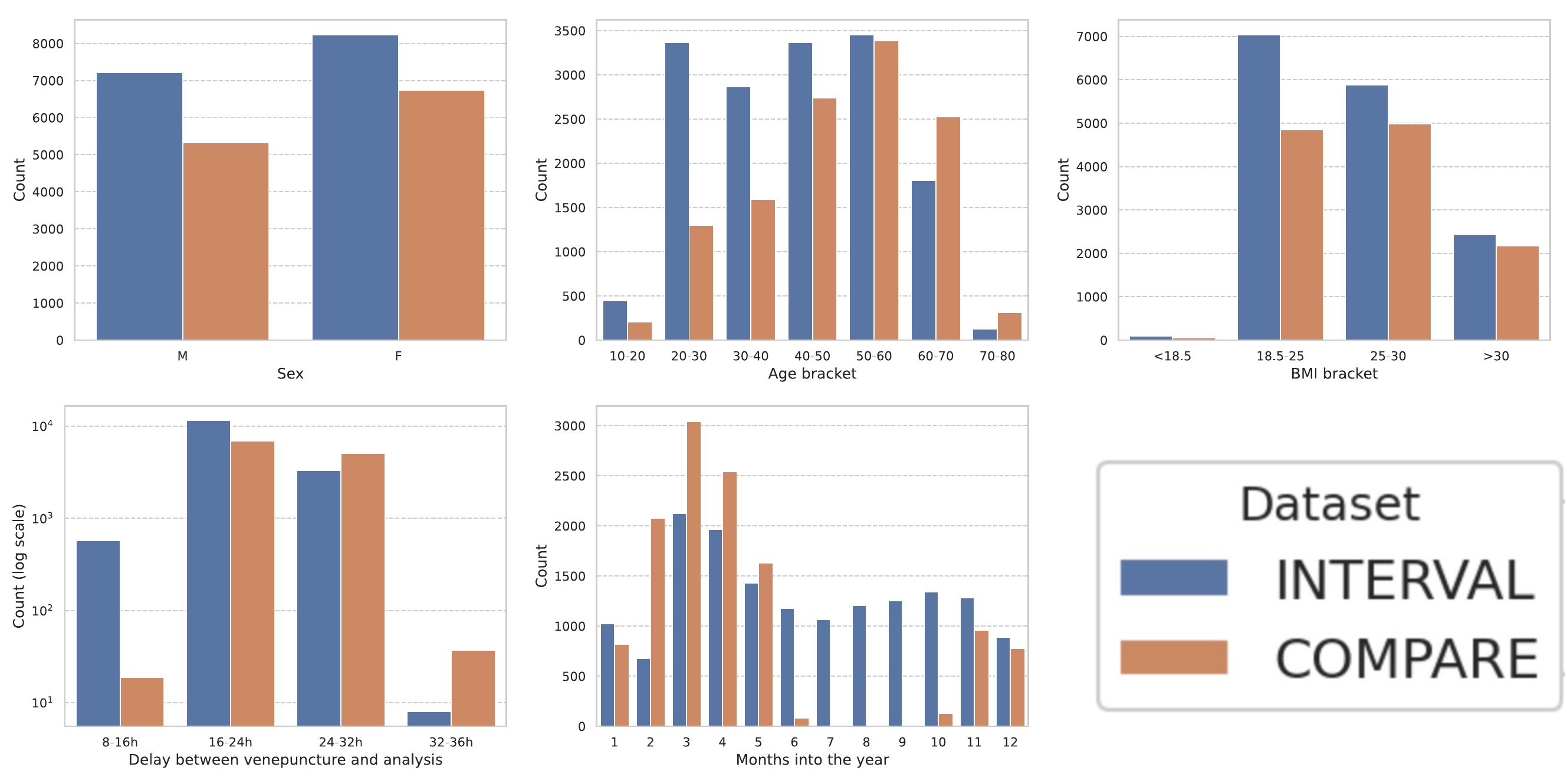}
\caption{\label{fig:INTERVAL_COMPARE_distributions}
Histogram plots of the various task and domain label distributions in the pre-filtered rich FBC data from the INTERVAL and COMPARE trials, coloured by blood donor study of origin.
}
\end{figure}

The pre-filtering on the available rich FBC data from the INTERVAL and COMPARE trials is outlined in Figure~\ref{fig:INTERVAL_COMPARE_flowchart}.
The distributions of the various task and domain labels in the pre-filtered dataset are displayed in Figure~\ref{fig:INTERVAL_COMPARE_distributions}.

\clearpage
\section{Preparation of FBC data from the Cambridge University Hospitals dataset}\label{sec:A_EpiCov_FBC_preprocess}
\begin{figure}[h]
\centering
\includegraphics[width=\linewidth]{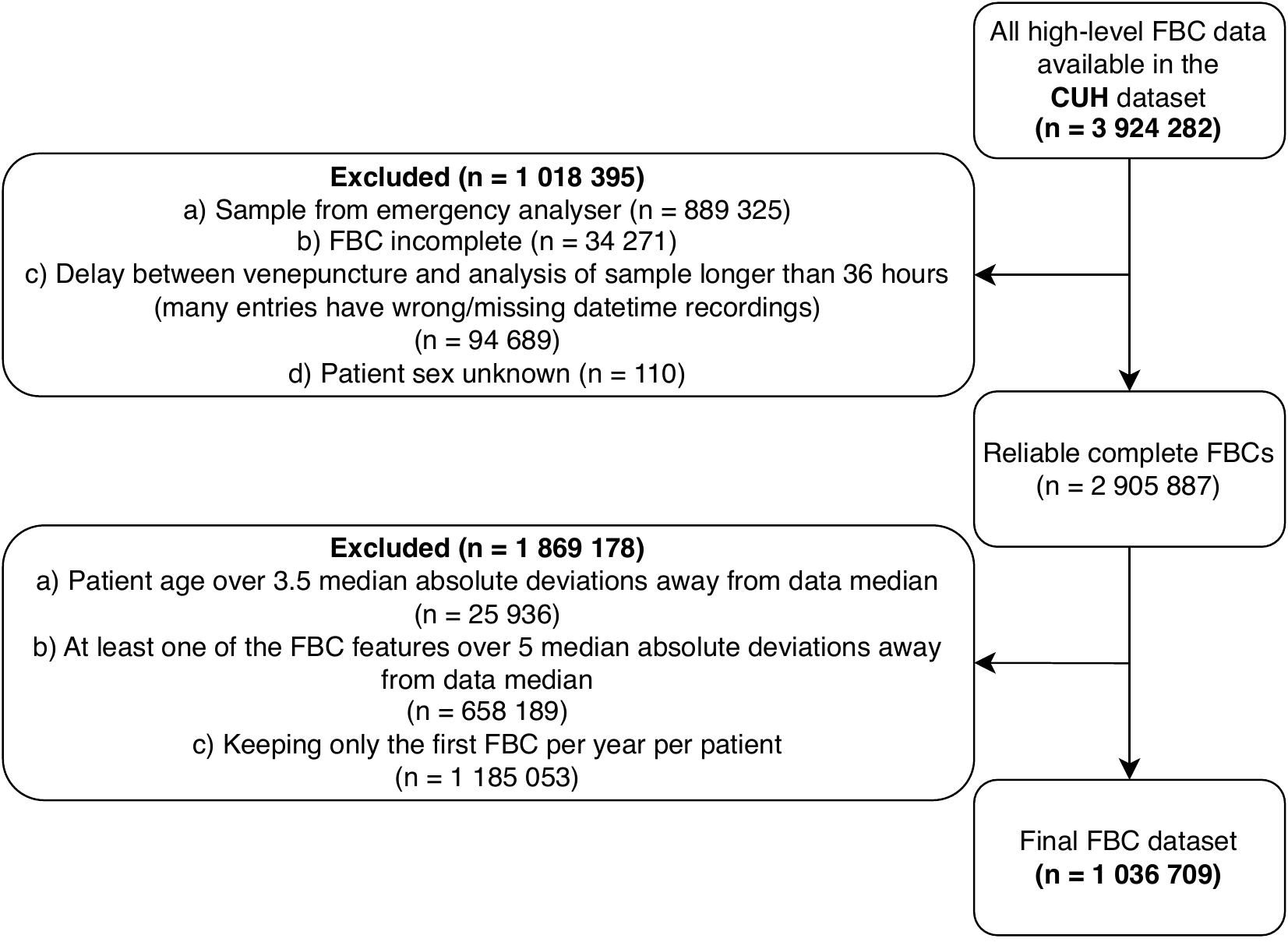}
\caption{\label{fig:EpiCov_flowchart}
Flowchart of the pre-filtering of the available high-level FBC data from the CUH dataset for the experiments in this paper. We chose to only include the first FBC per year per patient so as to not create a large bias towards very ill patients who would have many more FBCs in a short time interval than comparatively healthier ones.
}
\end{figure}

\begin{figure}[h]
\centering
\includegraphics[width=\linewidth]{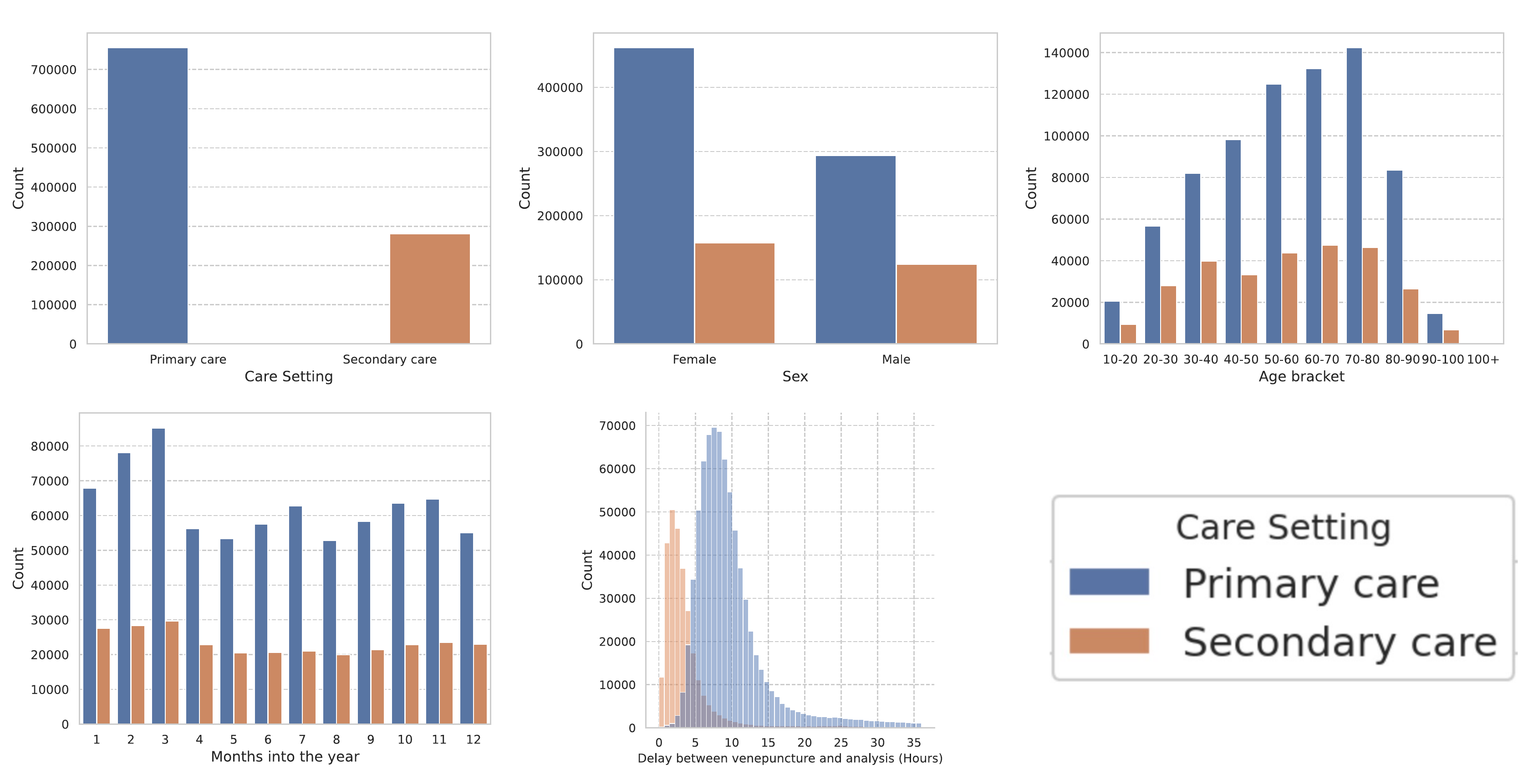}
\caption{\label{fig:EpiCov_distributions}
Histogram plots of the various task and domain label distributions in the pre-filtered high-level FBC data from the CUH dataset, coloured by patient care setting.
}
\end{figure}

The pre-filtering on the available high-level FBC data from the CUH dataset is outlined in Figure~\ref{fig:EpiCov_flowchart}.
The distributions of the various task and domain labels in the pre-filtered dataset are displayed in Figure~\ref{fig:EpiCov_distributions}.

\clearpage
\section{Supplementary: Dis-AE experiment results}\label{sec:A_Dis-AE_performance}
\subsection{Synthetic datasets experiments}\label{sec:A_Dis-AE_synthetic}
The quantitative performance for both task and domain objectives was measured using the classification accuracy of an XGBoost~\cite{chenXGBoostScalableTree2016} model and the variation $V^{\sup}$ on various representations of the data. Results for datasets A, B and C are displayed in Tables \ref{tab:accuracy_synthetic} and \ref{tab:variation_synthetic}.\\
It is visible that instance 5, being different from the other ones on different features, showed a drop in performance in all representations. Particularly the Dis-AE model seems to have difficulties as it is only trained to ignore differences in the features where instances 1 and 2 differ. This gives an early indication that the robustness of the model depends heavily on the domain diversity of its source data.\\
\begin{table}[h]
\centering
\begin{subtable}[t]{0.8\textwidth}
\begin{tabular}{c|c|c|ccc} \toprule
     & & \multicolumn{4}{c}{\textbf{Classification Accuracy}} \\
     Model & Task & Source & \multicolumn{3}{c}{Target} \\ \midrule
     & & Inst. 1\&2 & Inst. 3 & Inst. 4 & Inst. 5\\
    Raw data & Task 1 & 97 \% & 97 \% & 96 \% & \textbf{72 \%}\\
    Vanilla AE & Task 1 & 97 \% & 95 \% & 84 \% & 50 \%\\
    Dis-AE & Task 1 & \textbf{98 \%} & \textbf{98 \%} & \textbf{97 \%} &  53 \%\\\bottomrule
\end{tabular}
\caption{\footnotesize Dataset A}
\label{tab:table1_a}
\end{subtable}

\medskip 

\begin{subtable}[t]{0.8\textwidth}
\begin{tabular}{c|c|c|ccc} \toprule
     & & \multicolumn{4}{c}{\textbf{Classification Accuracy}} \\
     Model & Task & Source & \multicolumn{3}{c}{Target}  \\ \midrule
     & & Inst. 1\&2 & Inst. 3 & Inst. 4 & Inst. 5\\
    Raw data & Task 1 & 97 \% & 97 \% & 96 \% & \textbf{71 \%}\\
    Vanilla AE & Task 1 & 96 \% & 94 \% & 76 \% & 51 \%\\
    Dis-AE & Task 1 & \textbf{98 \%} & \textbf{97 \%} & \textbf{96 \%} &  53 \%\\\bottomrule
\end{tabular}
\caption{\footnotesize Dataset B}
\label{tab:table1_b}
\end{subtable}

\medskip 

\begin{subtable}[t]{0.8\textwidth}
\begin{tabular}{c|c|c|ccc} \toprule
 & & \multicolumn{4}{c}{\textbf{Classification Accuracy}} \\
 Model & Task & Source & \multicolumn{3}{c}{Target} \\ \midrule
 & & Inst. 1\&2 & Inst. 3 & Inst. 4 & Inst. 5\\
Raw data & Task 1 & \textbf{83 \%} & \textbf{80 \%} & 75 \% & 75 \%\\
 & Task 2 & 81 \% & 79 \% & 65 \% & \textbf{76 \%}\\
 & Task 3 & 74 \% & 68 \% & 66 \% & 67 \%\\
Vanilla AE & Task 1 & 68 \% & 69 \% & 68 \% & 64 \%\\
 & Task 2 & 73 \% & 73 \% & 73 \% & 68 \%\\
 & Task 3 & 67 \% & 63 \% & 67 \% & 63 \%\\
Dis-AE & Task 1 & 75 \% & 76 \% & 74 \% &  57 \%\\
 & Task 2 & 77 \% & 78 \% & \textbf{76 \%} & 70 \%\\
 & Task 3 & 70 \% & 69 \% & 67 \% & 62 \%\\
\bottomrule
\end{tabular}
\caption{\footnotesize Dataset C}
\label{tab:table1_c}
\end{subtable}
\caption{%
Accuracy of an XGBoost~\cite{chenXGBoostScalableTree2016} classifier on various representations of the synthetic datasets. The representations are the raw datasets, the embedding of a regular autoencoder and the embedding of our Dis-AE model. Target data performance is given for a classifier only trained on the full source data.}
\label{tab:accuracy_synthetic}
\end{table}

\begin{table}[h]
\centering
\begin{subtable}[t]{0.75\textwidth}
\begin{tabular}{c|c|ccc} \toprule
     & \multicolumn{4}{c}{\textbf{Variation $V^{\sup}$}} \\
     Model & Within source & \multicolumn{3}{c}{Source + target} \\ \midrule
     & Inst. 1\&2 & Inst. 3 & Inst. 4 & Inst. 5\\
    Raw data & 3.3 & 3.9 & 11.5 & 11.4\\
    Vanilla AE & 3.6 & 4.0 & 9.8 & 17.4\\
    Dis-AE & \textbf{0.01} & \textbf{0.04} & \textbf{0.1} &  \textbf{8.5}\\\bottomrule
\end{tabular}
\caption{\footnotesize Dataset A}
\label{tab:table1_a}
\end{subtable}

\medskip 

\begin{subtable}[t]{0.75\textwidth}
\begin{tabular}{c|c|ccc} \toprule
     & \multicolumn{4}{c}{\textbf{Variation $V^{\sup}$}} \\
     Model & Within source & \multicolumn{3}{c}{Source + target} \\ \midrule
     & Inst. 1\&2 & Inst. 3 & Inst. 4 & Inst. 5\\
    Raw data & 10.4 & 10.4 & 12.2 & 11.8\\
    Vanilla AE & 7.0 & 6.9 & 10.0 & 12.2\\
    Dis-AE & \textbf{0.1} & \textbf{0.1} & \textbf{0.3} &  \textbf{8.4}\\\bottomrule
\end{tabular}
\caption{\footnotesize Dataset B}
\label{tab:table1_b}
\end{subtable}

\medskip 

\begin{subtable}[t]{0.68\textwidth}
\begin{tabular}{c|c|ccc} \toprule
 & \multicolumn{4}{c}{\textbf{Variation $V^{\sup}$}} \\
 Model & Within source & \multicolumn{3}{c}{Source + target} \\ \midrule
 & Inst. 1\&2 & Inst. 3 & Inst. 4 & Inst. 5\\
Raw data & 10.4 & 10.4 & 12.2 & 11.8\\
Vanilla AE & 6.4 & 6.5 & 6.5 & 5.9\\
Dis-AE & \textbf{0.7} & \textbf{0.8} & \textbf{3.8} & \textbf{2.4}\\
\bottomrule
\end{tabular}
\caption{\footnotesize Dataset C}
\label{tab:table1_c}
\end{subtable}
\caption{%
Variation $V^{\sup}$ of various representations of the synthetic datasets. The representations are the raw datasets, the embedding of a regular autoencoder and the embedding of our Dis-AE model. $V^{\sup}$ gives a measure of the maximum domain shift in each representation. (We used $\rho_{\tx{W}_2}$ for the calculation of $V^{\sup}$ as the domain shift in the synthetic datasets is so large that $\rho_\tx{JSD}$ always returns 1 on the raw data and vanilla AE latent space.)
}
\label{tab:variation_synthetic}
\end{table}

\clearpage
\subsection{Rich FBC data experiments}\label{sec:A_Dis-AE_FBC_rich}
This appendix section displays further more detailed results on the per-class accuracies for the rich FBC data experiments. It is worth noting that all classes which have \SI{0}{\percent} classification accuracy across the board were only represented by very few data points (1-10).
\begin{table}[h]
\centering
\begin{subtable}[t]{0.68\textwidth}
\begin{tabular}{c|c|c|c} \toprule
     & & \multicolumn{2}{c}{\textbf{Class Accuracy}} \\
     Sex & Representation & Source & Target \\
    & & INTERVAL & COMPARE\\
    \hline
     & Raw data & 85 \% & 76 \%\\
    Male & Vanilla AE & 83 \% & 74 \%\\
    & Dis-AE & 83 \% & 70 \%\\
    \hline
    & Raw data & 89 \% & 93 \%\\
    Female & Vanilla AE & 87 \% & 92 \%\\
    & Dis-AE & 87 \% & 93 \%\\
    \bottomrule
\end{tabular}
\label{tab:table1_a}
\end{subtable}

\medskip 

\begin{subtable}[t]{0.68\textwidth}
\begin{tabular}{c|c|c|c} \toprule
     & & \multicolumn{2}{c}{\textbf{Class Accuracy}} \\
     BMI & Representation & Source & Target \\
    & & INTERVAL & COMPARE\\
    \hline
     & Raw data & 0 \% & 0 \%\\
    $\leq$ 18.5 & Vanilla AE & 0 \% & 0 \%\\
    & Dis-AE & 0 \% & 0 \%\\
    \hline
     & Raw data & 64 \% & 72 \%\\
    18.5-25 & Vanilla AE & 63 \% & 69 \%\\
    & Dis-AE & 62 \% & 67 \%\\
    \hline
     & Raw data & 53 \% & 41 \%\\
    25-30 & Vanilla AE & 51 \% & 45 \%\\
    & Dis-AE & 51 \% & 46 \%\\
    \hline
     & Raw data & 24 \% & 22 \%\\
    $\geq$ 30 & Vanilla AE & 21 \% & 18 \%\\
    & Dis-AE & 20 \% & 21 \%\\
    \bottomrule
\end{tabular}
\label{tab:table1_a}
\end{subtable}
\caption{%
Performance on rich FBC datasets (part 1/2).\\
Tasks: sex, age bracket, BMI range\\
Domains: haem. analyser, delay between venepuncture and analysis, time of year}
\label{tab:A_performance_FBCdata_full_part1}
\end{table}

\begin{table}[h]
\centering
\begin{subtable}[t]{0.68\textwidth}
\begin{tabular}{c|c|c|c} \toprule
     & & \multicolumn{2}{c}{\textbf{Class Accuracy}} \\
     Age & Representation & Source & Target \\
    & & INTERVAL & COMPARE\\
    \hline
     & Raw data & 1 \% & 1 \%\\
    $\leq$ 20 & Vanilla AE & 2 \% & 1 \%\\
    & Dis-AE & 0 \% & 1 \%\\
    \hline
     & Raw data & 35 \% & 52 \%\\
    20-30 & Vanilla AE & 32 \% & 53 \%\\
    & Dis-AE & 25 \% & 42 \%\\
    \hline
     & Raw data & 17 \% & 17 \%\\
    30-40 & Vanilla AE & 15 \% & 17 \%\\
    & Dis-AE & 14 \% & 16 \%\\
    \hline
     & Raw data & 30 \% & 25 \%\\
    40-50 & Vanilla AE & 29 \% & 21 \%\\
    & Dis-AE & 28 \% & 22 \%\\
    \hline
     & Raw data & 45 \% & 31 \%\\
    50-60 & Vanilla AE & 45 \% & 28 \%\\
    & Dis-AE & 45 \% & 34 \%\\
    \hline
     & Raw data & 30 \% & 20 \%\\
    60-70 & Vanilla AE & 27 \% & 20 \%\\
    & Dis-AE & 24 \% & 18 \%\\
    \hline
     & Raw data & 1 \% & 0 \%\\
    $\geq$ 70 & Vanilla AE & 0 \% & 0 \%\\
    & Dis-AE & 0 \% & 0 \%\\
    \bottomrule
\end{tabular}
\label{tab:table1_a}
\end{subtable}
\caption{%
Performance on rich FBC datasets (part 2/2).\\
Tasks: sex, age bracket, BMI range\\
Domains: haem. analyser, delay between venepuncture and analysis, time of year}
\label{tab:A_performance_FBCdata_full_part2}
\end{table}

\clearpage
\subsection{Cambridge University Hospitals data experiments}\label{sec:A_EpiCov_Results}
This appendix section includes experiment results from the CUH dataset experiments. The results are presented similarly to those of the other datasets and are included for sake of completeness. Interestingly, comparing Table~\ref{tab:selection_scores} and Table~\ref{tab:FBC_variations_EpiCov}, we see that the Dis-AE model struggles to achieve good generalisation and classification performance on the care setting task and the sample age domain. Indeed we found that sample age distributions between primary and secondary care patients barely overlapped and hence this task and domain are highly correlated. In trying to remove domain information for sample age the model inadvertently dampened the classification performance of the care setting task. Optimising this task and domain performance is non-trivial and requires careful tuning of the $\alpha$ and $\lambda$ hyperparameters in the model loss function.
\begin{figure}
     \centering
     \begin{subfigure}[b]{0.48\textwidth}
         \centering
         \includegraphics[width=\textwidth]{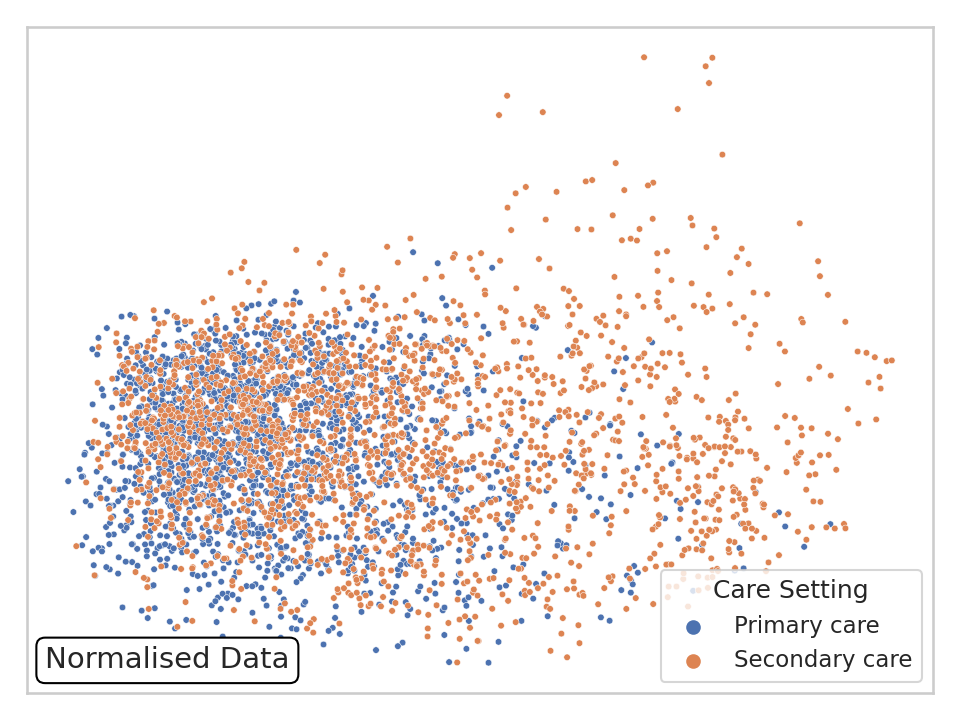}
         \caption{}
     \end{subfigure}
     \hfill
     \begin{subfigure}[b]{0.48\textwidth}
         \centering
         \includegraphics[width=\textwidth]{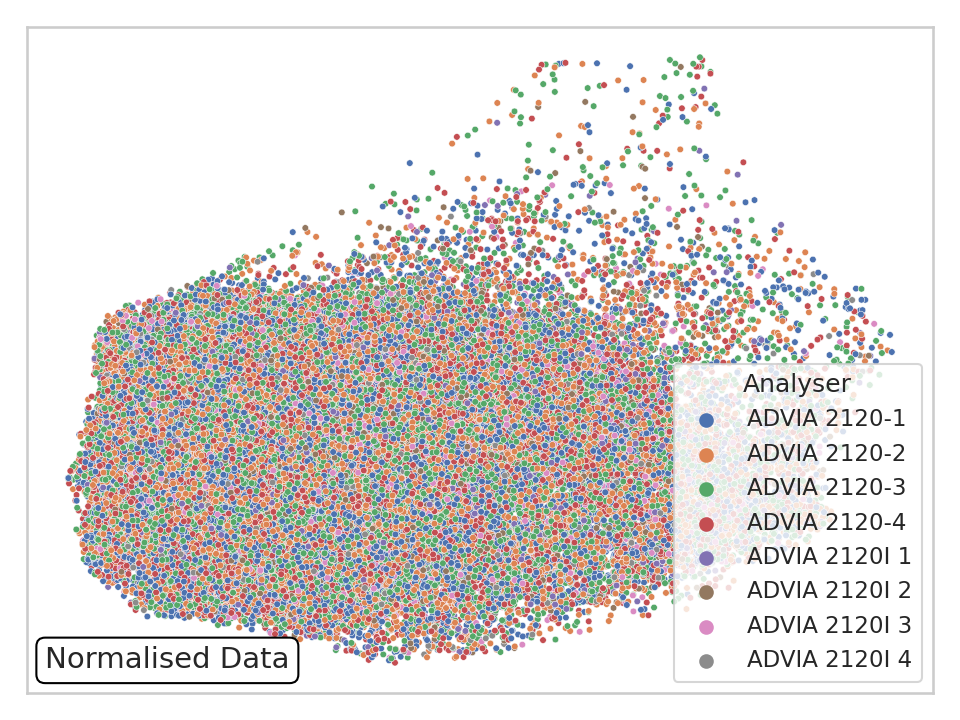}
         \caption{}
     \end{subfigure}\\
     \begin{subfigure}[b]{0.48\textwidth}
         \centering
         \includegraphics[width=\textwidth]{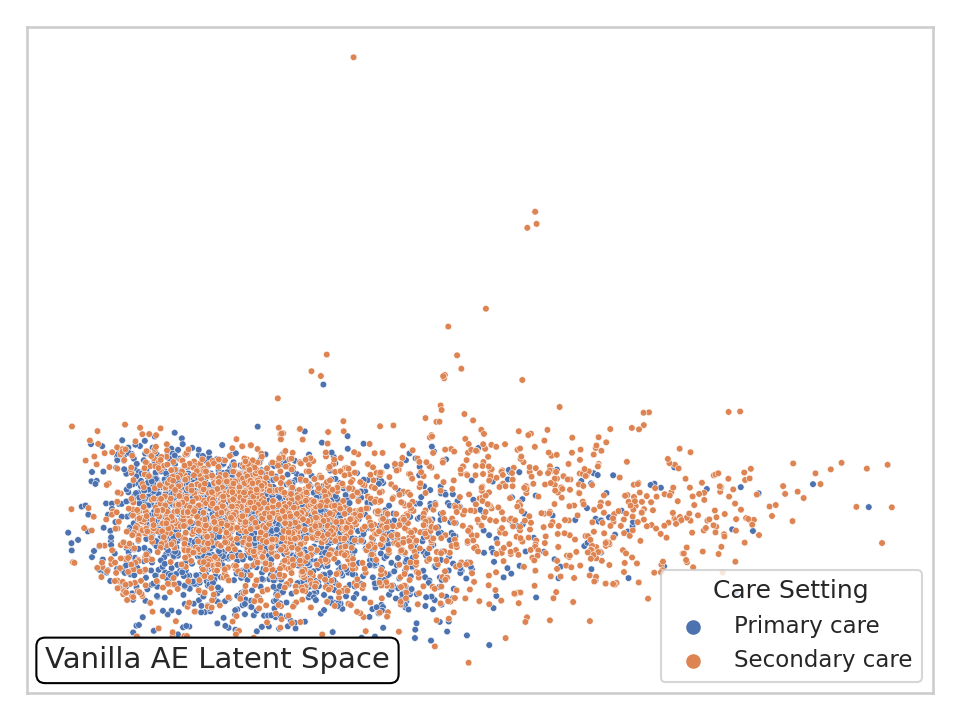}
         \caption{}
     \end{subfigure}
     \hfill
     \begin{subfigure}[b]{0.48\textwidth}
         \centering
         \includegraphics[width=\textwidth]{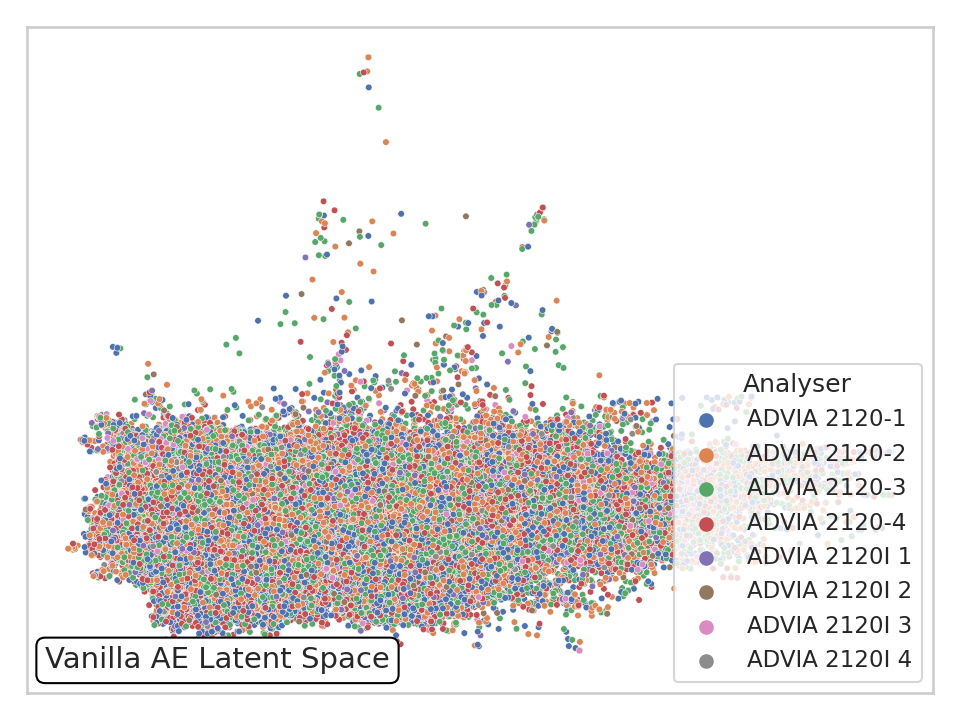}
         \caption{}
     \end{subfigure}\\
     \begin{subfigure}[b]{0.48\textwidth}
         \centering
         \includegraphics[width=\textwidth]{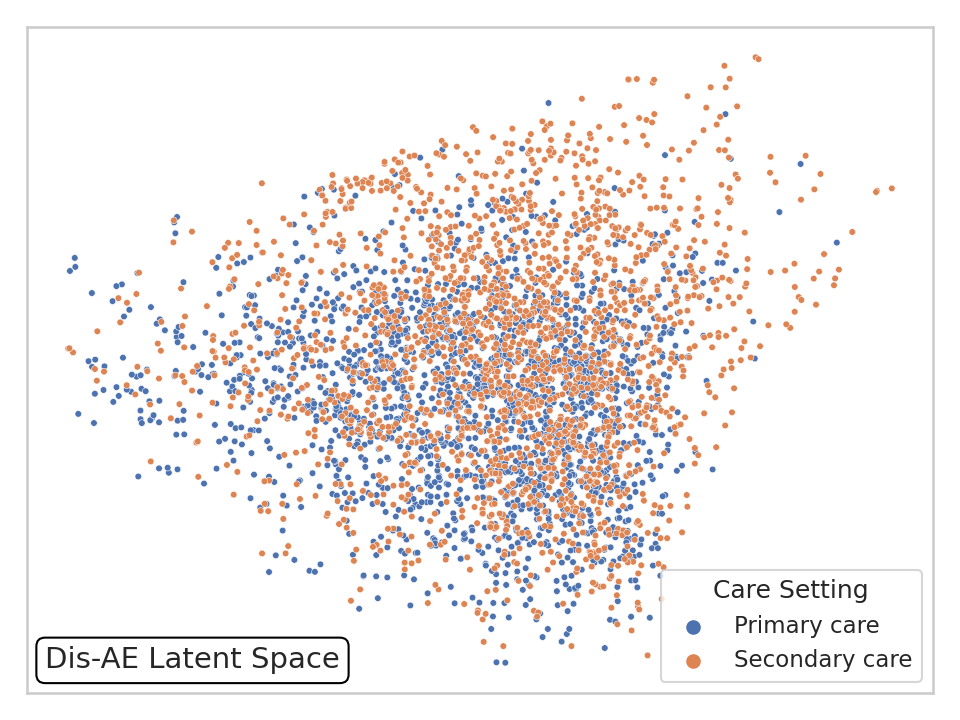}
         \caption{}
     \end{subfigure}
     \hfill
     \begin{subfigure}[b]{0.48\textwidth}
         \centering
         \includegraphics[width=\textwidth]{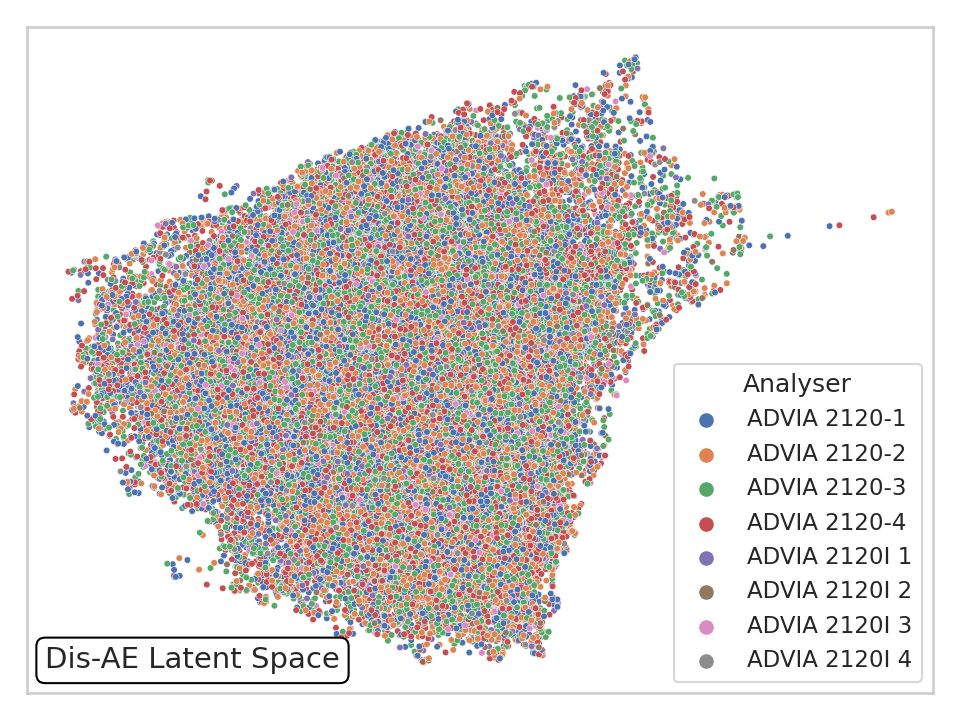}
         \caption{}
     \end{subfigure}
        \caption{HL-FBC hospital patient data PHATE~\cite{moonVisualizingStructureTransitions2019} visualisation of normalised raw data (row 1) and latent spaces of a vanilla AE (row 2) and the Dis-AE model (row 3). The images in the left column are coloured by patient care setting (example task), while those in the right column are coloured by haematology analyser (example domain). The machine learning models were only trained on data measured on the \texttt{ADVIA 2120I 2} and \texttt{ADVIA 2120I 3} analysers. In all plots, 2000 points per class were randomly selected from the calculated PHATE representation.}
        \label{fig:AEvDis-AE_FBC_PHATE_EpiCov}
\end{figure}
\begin{table}[h]
\begin{center}
\begin{tabularx}{0.78\textwidth}{l|c|c|c}
\toprule
     & & \multicolumn{2}{c}{\textbf{Variation $V^{\sup}$}} \\
    Representation & Domain & Within source & Source and target \\
     & & (2120I 2 \& 3) & (All analysers)\\
     \midrule
    & Analyser & 0.28 & 0.56\\
     (A) Raw data & Sample age & 0.64 & 0.49\\
     & Time of year & 0.19 & 0.83\\
     \hline
     & Analyser & 0.15 & 0.39\\
     (B) Vanilla AE & Sample age & 0.78 & 0.53\\
     & Time of year & 0.19 & \textbf{0.68}\\
     \hline
     & Analyser & \textbf{0.04} & \textbf{0.37}\\
     (C) Dis-AE & Sample age & \textbf{0.64} & \textbf{0.46}\\
     & Time of year & \textbf{0.17} & 0.71\\
    \bottomrule
\end{tabularx}
\caption{Variation of HL-FBC data from the Cambridge University Hospitals patient dataset in different representations. The three representations are: (A) the unmodified data, (B) the latent space of a vanilla AE and (C) the latent space of the Dis-AE model. The variation $V^{\sup}$, calculated using $\rho_\tx{JSD}$, is shown per domain to show each covariate's contribution to the data's domain shift.}
\label{tab:FBC_variations_EpiCov}
\end{center}
\end{table}

\begin{table}[h]
\centering
\begin{subtable}[t]{0.68\textwidth}
\begin{tabular}{c|c|c|c} \toprule
     & & \multicolumn{2}{c}{\textbf{Class Accuracy}} \\
     Sex & Representation & Source & Target \\
    & & 2120I 2 \& 3 & (all other analysers)\\
    \hline
     & Raw data & 64 \% & 65 \%\\
    Male & Vanilla AE & 63 \% & 64 \%\\
    & Dis-AE & 63 \% & 65 \%\\
    \hline
    & Raw data & 83 \% & 83 \%\\
    Female & Vanilla AE & 83 \% & 83 \%\\
    & Dis-AE & 83 \% & 83 \%\\
    \bottomrule
\end{tabular}
\label{tab:table1_a}
\end{subtable}

\medskip 

\begin{subtable}[t]{0.68\textwidth}
\begin{tabular}{c|c|c|c} \toprule
     & & \multicolumn{2}{c}{\textbf{Class Accuracy}} \\
     Care Setting & Representation & Source & Target \\
    & & 2120I 2 \& 3 & (all other analysers)\\
    \hline
     & Raw data & 90 \% & 92 \%\\
    Primary & Vanilla AE & 90 \% & 91 \%\\
    & Dis-AE & 90 \% & 91 \%\\
    \hline
    & Raw data & 53 \% & 46 \%\\
    Secondary & Vanilla AE & 52 \% & 46 \%\\
    & Dis-AE & 41 \% & 38 \%\\
    \bottomrule
\end{tabular}
\label{tab:table1_a}
\end{subtable}
\caption{%
Performance on HL-FBC dataset (part 1/2).\\
Tasks: sex, age bracket, care setting\\
Domains: haem. analyser, delay between venepuncture and analysis, time of year}
\label{tab:A_performance_FBCdata_full_part1}
\end{table}

\begin{table}[h]
\centering
\begin{subtable}[t]{0.68\textwidth}
\begin{tabular}{c|c|c|c} \toprule
     & & \multicolumn{2}{c}{\textbf{Class Accuracy}} \\
     Age & Representation & Source & Target \\
    & & 2120I 2 \& 3 & (all other analysers)\\
    \hline
     & Raw data & 3 \% & 2 \%\\
    $\leq$ 20 & Vanilla AE & 3 \% & 2 \%\\
    & Dis-AE & 3 \% & 2 \%\\
    \hline
     & Raw data & 9 \% & 5 \%\\
    20-30 & Vanilla AE & 9 \% & 5 \%\\
    & Dis-AE & 7 \% & 5 \%\\
    \hline
     & Raw data & 28 \% & 21 \%\\
    30-40 & Vanilla AE & 27 \% & 20 \%\\
    & Dis-AE & 26 \% & 21 \%\\
    \hline
     & Raw data & 11 \% & 10 \%\\
    40-50 & Vanilla AE & 11 \% & 9 \%\\
    & Dis-AE & 10 \% & 9 \%\\
    \hline
     & Raw data & 28 \% & 30 \%\\
    50-60 & Vanilla AE & 27 \% & 30 \%\\
    & Dis-AE & 30 \% & 30 \%\\
    \hline
     & Raw data & 19 \% & 21 \%\\
    60-70 & Vanilla AE & 18 \% & 21 \%\\
    & Dis-AE & 19 \% & 20 \%\\
    \hline
     & Raw data & 43 \% & 42 \%\\
    70-80 & Vanilla AE & 43 \% & 41 \%\\
    & Dis-AE & 44 \% & 42 \%\\
    \hline
     & Raw data & 27 \% & 30 \%\\
    80-90 & Vanilla AE & 27 \% & 30 \%\\
    & Dis-AE & 27 \% & 29 \%\\
    \hline
     & Raw data & 1 \% & 1 \%\\
    90-100 & Vanilla AE & 1 \% & 1 \%\\
    & Dis-AE & 1 \% & 1 \%\\
    \hline
     & Raw data & 0 \% & 0 \%\\
    $\geq$ 100 & Vanilla AE & 0 \% & 0 \%\\
    & Dis-AE & 0 \% & 0 \%\\
    \bottomrule
\end{tabular}
\label{tab:table1_a}
\end{subtable}
\caption{%
Performance on HL-FBC dataset (part 2/2).\\
Tasks: sex, age bracket, care setting\\
Domains: haem. analyser, delay between venepuncture and analysis, time of year}
\label{tab:A_performance_FBCdata_full_part2}
\end{table}

\end{appendices}

\end{document}